\documentclass[journal]{IEEEtran}
\usepackage{graphicx,amssymb,soul,color,graphics,epsfig,mathptmx,times,amsmath,amssymb,color,subfigure,soul,moreverb,ifpdf,cancel,caption,xcolor,amsthm,framed,subfigure,dblfloatfix}

\usepackage{cite}
\colorlet{shadecolor}{yellow!100}
\ifCLASSINFOpdf
\fi
\UseRawInputEncoding
\begin{document}
\title{Towards Agrobots: Trajectory Control of an Autonomous Tractor Using Type-2 Fuzzy Logic Controllers}
\author{Erdal~Kayacan,~\IEEEmembership{Senior Member, IEEE,}~Erkan~Kayacan,~\IEEEmembership{Student Member, IEEE,}~Herman~Ramon,~\IEEEmembership{}~Okyay~Kaynak,~\IEEEmembership{Fellow Member, IEEE,} and Wouter~Saeys~\IEEEmembership{}
\thanks{E. Kayacan, E. Kayacan, H. Ramon and W. Saeys are with the Division of Mechatronics, Biostatistics and Sensors, Department of Biosystems, University of Leuven (KU Leuven), Kasteelpark Arenberg 30, B-3001 Leuven, Belgium. e-mail: {\tt\small \{erdal.kayacan, erkan.kayacan, herman.ramon, wouter.saeys\}@biw.kuleuven.be} }
\thanks{O. Kaynak is with the Department of Electrical and Electronics Engineering, Bogazici University, 34342, Istanbul, Turkey. e-mail: {\tt\small okyay.kaynak@boun.edu.tr}}}
\markboth{\textbf{PREPRINT VERSION:} IEEE /ASME Transactions on Mechatronics, , vol. 20, no. 1, pp. 287-298, Feb. 2015.}
{Shell \MakeLowercase{\textit{et al.}}: Bare Demo of IEEEtran.cls for Journals}
\maketitle

\begin{abstract}
Provision of some autonomous functions to an agricultural vehicle would lighten the job of the operator but in doing so, accuracy should not be lost to still obtain an optimal yield. Autonomous navigation of an agricultural vehicle involves the control of different dynamic subsystems, such as the yaw angle dynamics and the longitudinal speed dynamics.  In this study, a PID controller is used to control the longitudinal velocity of the tractor. For the control of the yaw angle dynamics, a PD controller works in parallel with a type-2 fuzzy neural network. In such an arrangement, the former ensures the stability of the related subsystem, while the latter learns the system dynamics and becomes the leading controller. In this way, instead of modeling the interactions between the subsystems prior to the design of a model-based control, we develop a control algorithm which learns the interactions on-line from the measured feedback error. In addition to the control of the stated subsystems, a kinematic controller is needed to correct the errors in both the $x$ and the $y$ axis for the trajectory tracking problem of the tractor. To demonstrate the real-time abilities of the proposed control scheme, an autonomous tractor is equipped with the use of reasonably priced sensors and actuators. Experimental results show the efficacy and the efficiency of the proposed learning algorithm.
\end{abstract}

\IEEEpeerreviewmaketitle
\section{Introduction}
\IEEEPARstart{I}{n}  recent years, considerable efforts are being made to automate functions in agricultural machinery and  production machines  that are still carried out  by humans, through the use of self-learning controllers and continuously adapt the machine settings to the crop/animal variability and the environmental conditions.

Agricultural production machines with higher efficiencies will be very important in the future because of the limited agricultural areas in the world and constantly rising energy and labor costs. One way to offset the rising labor costs may be to increase the capacity  by increasing the size of the machines. However, the limits in this direction will soon be reached as there is a maximum size that will still permit road transportation. On the other hand, the requirement for the minimization of the energy costs may make the multi-objective optimization problem in hand a difficult one to solve. A different approach to increase effectiveness  would be to increase the operational efficiency by the use of advanced learning algorithms, which can learn the  operational dynamics online and adjust the operational parameters accordingly. The motivation behind the use of self-learning controllers instead of conventional controllers for the control of agricultural production machines is that there are different subsystems interacting with each other in these machines, and well tuning of the controller coefficients simultaneously is a difficult task. Even if the operator becomes proficient in proper adjustment of the different controller coefficients, crop/animal variability and the environmental conditions force the operator to change the machine settings continuously resulting in the fact that adaptability is a must. In this study, the trajectory control of an autonomous tractor is considered and, for the provision of such adaptability, a Takagi-Sugeno-Kang (TSK) type-2 fuzzy neural network (T2FNN) with a sliding mode control (SMC) theory-based learning algorithm is proposed. Various uncertainties, disturbances and nonlinearities that would inherently exist in such a system are thus handled.

As for the structure, the combination of a conventional controller and an intelligent controller is chosen. The former is a proportional-derivative (PD) controller and the latter is a T2FNN in this study. Such a structure is called \emph{feedback error learning} (FEL) in literature; it was originally proposed in \cite{Kawato} for robot control in which a neural network based controller works  with a PD controller. In this approach, the output of the conventional controller is used as the learning error signal to train the intelligent controller. A recent extension of this approach to fuzzy neural networks (FNNs) can be seen in \cite{ascc09}. This study, presents a further extension by using SMC theory to train T2FNNs. The novelty of the approach is that instead of trying to minimize an error function, the learning parameters are tuned by the proposed algorithm in a way to enforce the error to satisfy a stable equation. The parameter update rules of T2FNNs are derived and the stability of the learning algorithm in the Lyapunov sense is proved.

For the training of a FNN or a T2FNN, two methods are widely used in literature. One of them is gradient-descent-based \cite{gradient} and therefore requires the computation of partial derivatives or sensitivity functions. It can be considered as an extension of the commonly used learning algorithm with back propagation of the error. There are some drawbacks of this method such as slow speed of learning, long computational time, and difficulties are met in determining the convergence and the stability of the learning scheme in an analytical way. What is more, as is the case with all repetitive algorithms, a number of numerical robustness issues may emerge when the algorithm is run over a long period of time \cite{Astrom_Witternmark}. In addition to these drawbacks, the tuning process can easily be trapped into a local minimum \cite{Venelinov_1}. Another well known learning algorithm is based on evolutionary computations with genetic algorithms \cite{celikyilmaz,genetik2}. Since genetic algorithms-based methods  basically do a random search, they are slower than gradient-descent algorithms and computationally more intensive. Moreover, the stability of such approaches is questionable and the optimal values for the stochastic operators are difficult to derive. In order to overcome these difficulties, SMC theory-based algorithms have been proposed in \cite{Parma,Yu} for the parameter update rules of artificial neural networks (ANNs) and FNNs as robust learning algorithms. Even though the main strength of SMC is its robustness, there are some disadvantages of it as well. For instance, when the system dynamics are close to the sliding surface, high frequency oscillations in the control input, also known as chattering, occur. One of the most common methods to eliminate the chattering is inserting a boundary layer to replace the corrective control by an equivalent one when the system is inside this layer \cite{Slotin_1}.

SMC is an approach that guarantees the robustness of a system in the case of external disturbances, parameter variations and uncertainties and as such has attracted the attention of many researchers to guarantee robustness in computationally intelligent architectures \cite{okyayhoca_survey1,okyayhoca_survey2}. The main idea behind this control scheme is to restrict the motion of the system in a plane referred to as the \emph{sliding surface}, where the predefined function of the error is zero \cite{Oniz}. SMC-based learning algorithms can, not only make the overall system more robust, but also ensure faster convergence than the traditional learning techniques in online tuning of ANNs and type-1 fuzzy neural networks (T1FNNs). There are various studies in literature that aim to use the robustness property of SMC in the learning process of ANNs and T1FNNs \cite{Efe2000}. Conversely, the robustness and the stability properties of soft computing-based control strategies can also be analyzed through the use of SMC theory \cite{Byungkook}.

The most common tools used in the literature to implement model free designs are ANNs and fuzzy logic systems (FLSs). On the fuzzy logic theory, Zadeh argues that \emph{"fuzzy logic is a precise conceptual system of reasoning, deduction and computation in which the objects of discourse and analysis are, or are allowed to be, associated with imperfect information. Imperfect information is information which in one or more respects is imprecise, uncertain, incomplete, unreliable, vague or partially true"} \cite{Zadehh}. ANNs are well known for their representation capability, even in the case of highly nonlinear systems. FNNs combine the advantages of both techniques, \emph{i.e.} the fuzzy reasoning ability of FLSs and the learning ability of ANNs.

The use of type-2 fuzzy sets has an advantage when it is difficult to determine the place of the membership functions (MFs) precisely \cite{Liang1}. Since there are infinite number of type-1 MFs within an interval in a type-2 MF, type-2 fuzzy logic systems (T2FLSs) appear to be a more promising method than their type-1 counterparts for handling uncertainties such as noisy data and changing environments \cite{chia09,Juang1}. In \cite{sepu06,Roo10}, the effects of the measurement noise in type-1 and type-2 fuzzy logic controllers and identifiers were simulated to perform a comparative analysis. It was concluded that the use of a type-2 fuzzy logic controller can be a better option than the use of its type-1 counterpart. These claims have been verified in a real-time application in \cite{MendelIE}. Even if there exists a number of papers in literature that claim that the performance of T2FLSs is better than T1FLSs under noisy conditions, this claim is tried to be justified by simulation or real-time studies only for some specific systems. However, in \cite{ellipsoidal}, a simpler T2FLS is considered with a novel MF in which the effect of input noise in the rule base is shown numerically in a general way, and it is concluded that T2FLSs should be used when needed, i.e. in the presence of noise and uncertainties in the system.

Whereas the secondary MFs can take values in the interval of [0,1] in generalized T2FLSs, they are uniform functions that only take on values of 1 in interval T2FLSs. Since the general T2FLSs are computationally very demanding (this is because the type-reduction is computationally expensive), the use of interval T2FLSs is more commonly seen in literature. In an interval TSK T2FLS, there exists a design parameter that weights the sharing of lower and upper firing levels of each fired rule. That parameter can either be fixed or optimized online. While the least mean square method is generally used to find the optimal value of this parameter, an SMC theory-based learning algorithm is proposed for tuning it in \cite{WCCI2012_australia}.

The contributions of this paper are as follows:
\begin{itemize}
  \item An SMC theory-based learning algorithm is proposed for the parameter tuning of T2FNNs, including the design parameter that weights the sharing of the lower and the upper firing levels of each fired rule, and its stability in the Lyapunov sense  is proved.
  \item The proposed learning algorithm is tested on the trajectory tracking problem of an autonomous agricultural tractor in the presence of various nonlinearities and uncertainties in real-time.
  \item A practical mechatronic system, illustrating how control, sensing and actuation can be integrated to achieve an intelligent system, is designed and presented.
\end{itemize}

The body of the paper contains five sections: In Section II, the dynamic equations of the autonomous tractor are presented. In Section III, the overall control is scheme is given. The proposed sliding mode FEL approach is presented and the parameter update rules for T2FNNs are proposed for the case of triangular MFs in Section IV. In Section V, real-time results are given. Finally, conclusions are presented in Section VI.

\section{Mathematical Description of the Tractor}

\subsection{Kinematic Model}
\begin{figure}[h!]
\centering
  \includegraphics[width=2.5in]{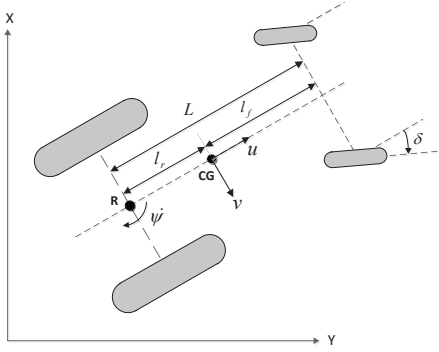}\\
  \caption{The autonomous tractor}
  \label{kinematic}
\end{figure}

The schematic diagram of the autonomous tractor is presented in Fig. \ref{kinematic}.

The linear velocities $\dot{x}$, $\dot{y}$ and the yaw rate $\dot{\psi}$ at the $R$ point are written as follows:
\begin{eqnarray}\label{kinematicmodeltractor}
\left[
  \begin{array}{c}
   \dot{x} \\
   \dot{y} \\
   \dot{\psi} \\
  \end{array}
  \right]
  =
  \left[
  \begin{array}{c}
   {u} \cos{\psi} \\
   {u} \sin{\psi} \\
   \frac{u tan{\delta}}{L} \\
  \end{array}
  \right]
\end{eqnarray}
where $u$, $\psi$, $\delta$ and $L$ represent the longitudinal velocity, the yaw angle defined on the point $R$ on the tractor, the steering angle of the front wheel, the distance between the front and the rear axles of the tractor, respectively.

Considering the center of gravity (CG) shown in Fig. \ref{kinematic}, the linear velocities $\dot{x}$, $\dot{y}$ and the yaw rate $\dot{\psi}$ of CG  can be written as follows:
\begin{eqnarray}\label{kinematicmodeltractor2}
\left[
  \begin{array}{c}
   \dot{x} \\
   \dot{y} \\
   \dot{\psi} \\
  \end{array}
  \right]
  =
  \left[
  \begin{array}{c}
   {u} \cos{\psi} - v \sin{\psi} \\
   {u} \sin{\psi} + v \cos{\psi} \\
   \frac{u tan{\delta}}{L} \\
  \end{array}
  \right]
\end{eqnarray}
where $v$ equals to the multiplication of $\dot{\psi}$ and $l_r$.

\subsection{The Yaw Dynamics Model}
\begin{figure}[h!]
\centering
\subfigure[ ]{
\includegraphics[width=1.5in]{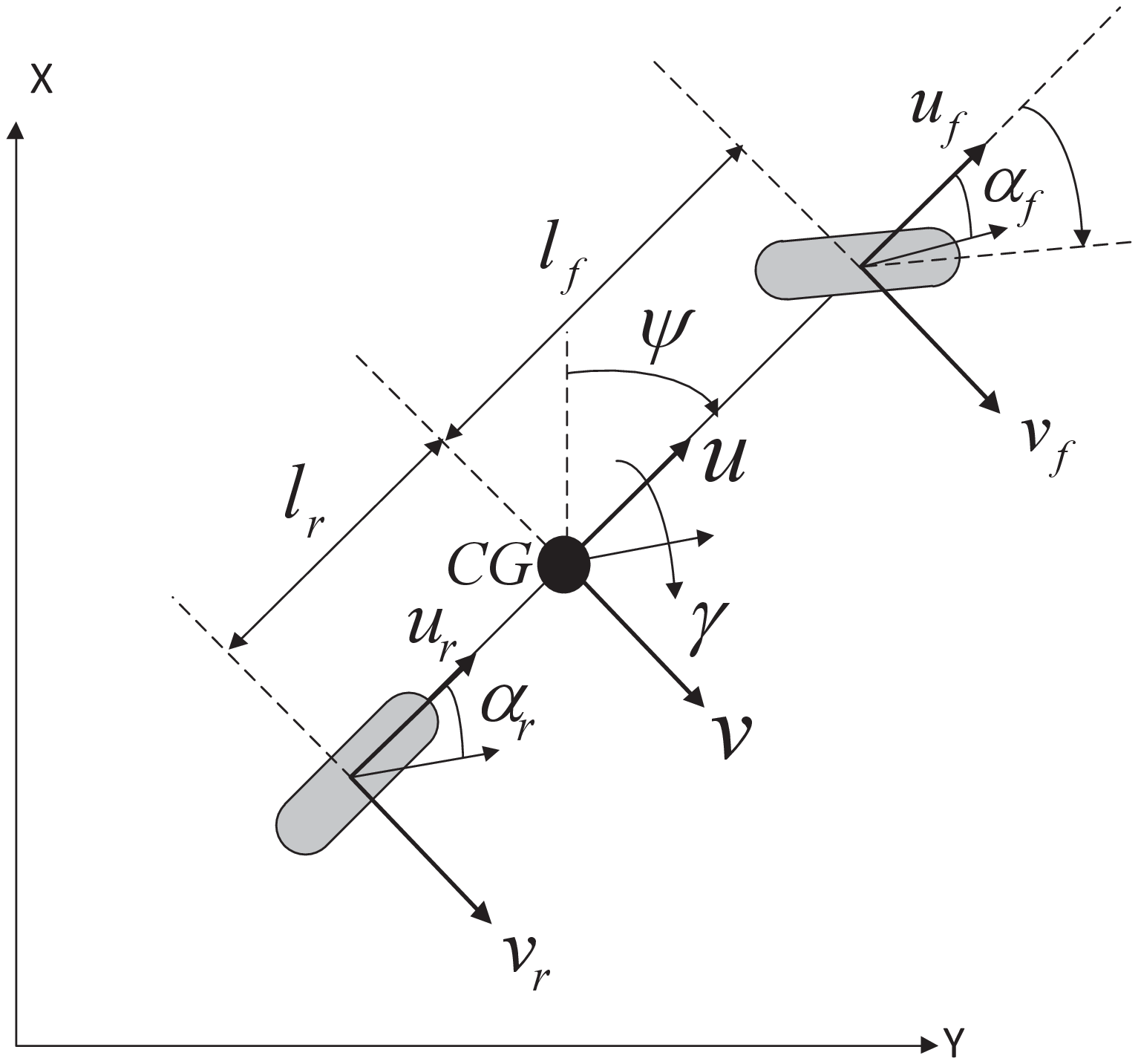}
\label{angle}
}
\subfigure[ ]{
\includegraphics[width=1.5in]{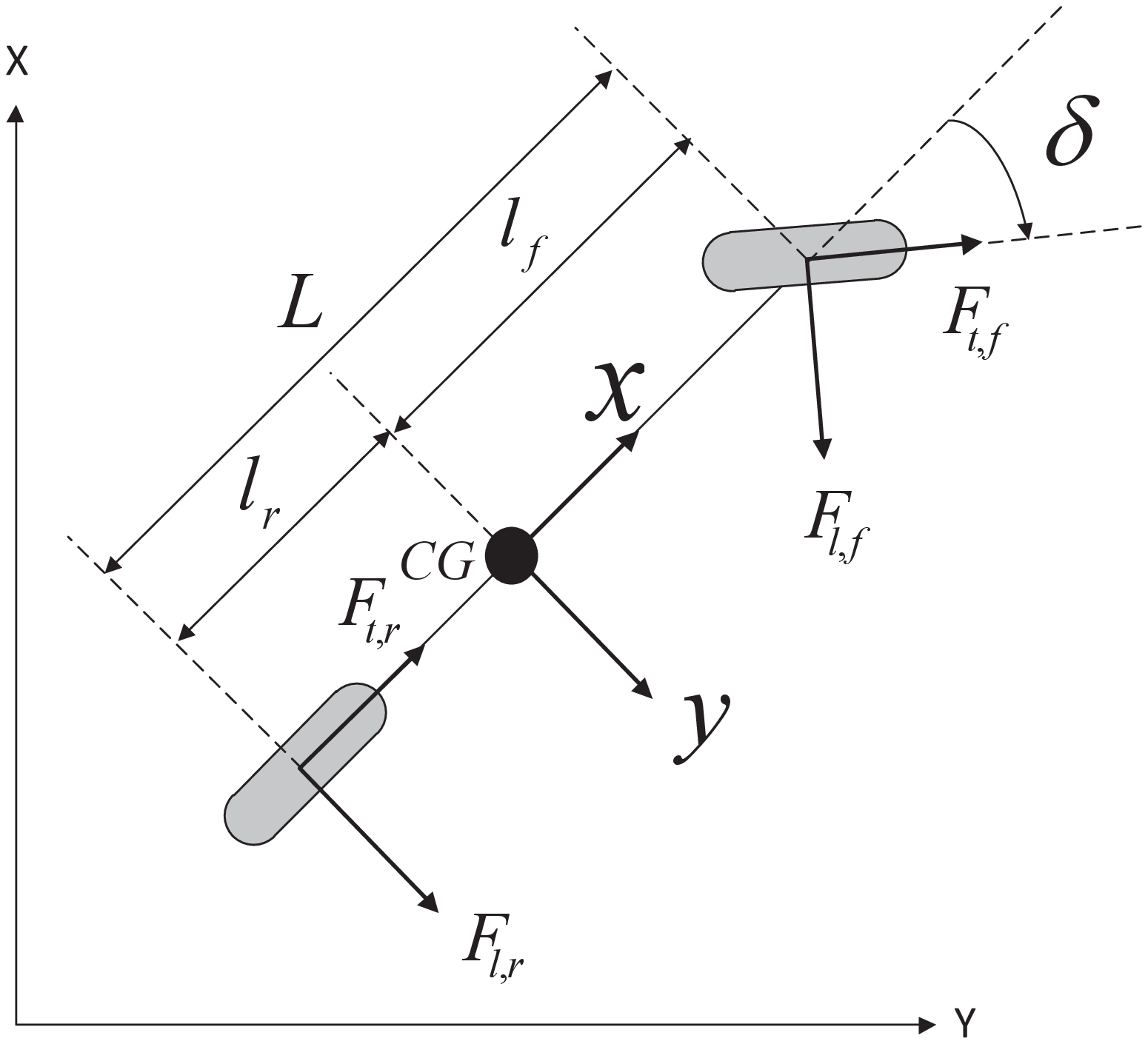}
\label{force}
}
\label{angleforce}
\caption[Optional caption for list of figures]{Dynamic bicycle model for a tractor: (a) velocities and side-slip angles (b) forces on the rigid body of the system}
\vskip -0.1cm
\end{figure}

The velocities and the side-slip angles on the rigid body of the tractor are presented in Fig. \ref{angle}. Similarly, the forces on the rigid body of the tractor are shown in Fig. \ref{force}.

The lateral dynamics of the tractor can be written as follows:
\begin{equation}\label{lateralmotionoftractor}
   m (\dot{v} + u \gamma ) = F _{t,f} \sin\delta + F _{l,f} \cos\delta + F _{l,r}
\end{equation}
where $m$, $v$, $u$, $\gamma$, $F _{t,f}$, $F _{l,f}$, $F _{l,r}$ and $\delta$ represent the mass of the tractor, the lateral velocity of the  CG,  the longitudinal velocity of the CG, the yaw rate, the traction and the lateral forces on the front wheel, the lateral force on the rear wheel and the steering angle of the front wheel, respectively.

The yaw dynamics of the tractor are written as follows:
\begin{eqnarray}\label{yawmotionoftractor}
   I _{z} \dot{\gamma}  & =  & l _{f} ( F _{t,f} \sin\delta + F _{l,f} \cos\delta ) - l _{r} F _{l,r}
\end{eqnarray}
where $l _{f}, l _{r}$ and $I_{z}$ respectively represent the distance between the front axle and the CG of the tractor, the distance between the rear axle and the CG of the tractor, and the inertial moment of the tractor.

The tire side-slip angles must be calculated in order to determine the forces caused by the slip. It is assumed that the steering angle of the front wheel is small, and this allows to make the following approximations: $\sin\delta\approx\delta$ and $\cos\delta\approx1$. The side-slip angles of the front ($\alpha _{f}$) and the rear tires ($\alpha _{r}$)  are written as follows:
\begin{equation}\label{sideslipangles}
   \alpha _{f}   =  \frac{v +  l_f \gamma}{u} - \delta \;\; \text{and} \;\; \alpha _{r}   =  \frac{v -  l_r \gamma}{u}
\end{equation}

To determine the lateral force on the tire, there are many different approaches in literature. In this study, the lateral tire forces are calculated using a linear model which assumes these to be proportional to the side-slip angles in \cite{Karkee,Piyabongkarn,Geng}
\begin{equation}\label{lateralforces}
    F _{l,i} = - C _{\alpha,i}  \alpha _{i}  \quad \quad i=\{f,r\}
\end{equation}
where $C _{\alpha,i}$, $i=\{f,r\}$, represents the cornering stiffness of the tires of the tractor. The tire cornering stiffness parameters are the averaged slopes of the lateral force characteristics in this method.

The equations of yaw motion of the autonomous tractor are written in state space form by combining \eqref{lateralmotionoftractor}, \eqref{yawmotionoftractor}, \eqref{sideslipangles} and \eqref{lateralforces} as follows:
\begin{eqnarray}\label{sondenklem}
\left[
  \begin{array}{c}
  \dot{v} \\
  \dot{\gamma} \\
  \end{array}
  \right]
=
\left[
  \begin{array}{ccc}
  A_{11} & A_{12}  \\
  A_{21}  & A_{22}  \\
  \end{array}
  \right]
\left[
  \begin{array}{c}
   v  \\
   \gamma  \\
  \end{array}
  \right]
  +
\left[
  \begin{array}{c}
  B_{1}  \\
  B_{2}  \\
   \end{array}
  \right]
\delta(t)
\end{eqnarray}
where
\begin{eqnarray}
   A_{11}&=& -\frac{C _{\alpha,f}+C _{\alpha,r}}{m u }, \nonumber \\
   A_{12}&=& \frac{-l_f C _{\alpha,f} + l_r C _{\alpha,r}}{m u } - u ,  \nonumber \\
   A_{21}&=&\frac{-l_f C _{\alpha,f} + l_r C _{\alpha,r}}{I_z u }, \nonumber \\
   A_{22}&=& - \frac{l_f^2 C _{\alpha,f} + l_r^2 C _{\alpha,r}}{I _{z} u }, \nonumber \\
   B_{1}&=& \frac{C _{\alpha,f}}{m}   \;\; , \;\;   B_{2}= \frac{l_f C _{\alpha,f}}{I _{z}}
\end{eqnarray} 
\section{Overall Control Scheme}
\subsection{Kinematic Controller}
The kinematic model is re-written in a state-space as follows:
\begin{eqnarray}\label{kinematicmodeltractorcontrol}
\left[
  \begin{array}{c}
   \dot{x} \\
   \dot{y} \\
   \dot{\psi} \\
  \end{array}
  \right]
  =
  \left[
  \begin{array}{cc}
  \cos{\psi} & - l_{r} \sin{\psi} \\
  \sin{\psi} &  l_{r} \cos{\psi} \\
   0 & 1\\
  \end{array}
  \right]
    \left[
  \begin{array}{c}
   {u} \\
   \gamma \\
  \end{array}
  \right]
\end{eqnarray}
where lateral velocity $v$ equals to $\gamma l_{r} $.

An inverse kinematic model is needed to calculate the reference speed and the yaw rate for the tractor. It is written as follows:
\begin{eqnarray}\label{inversekinematicmodeltractor}
\left[
  \begin{array}{c}
  u \\
  \gamma \\
  \end{array}
  \right]
  =
  \left[
  \begin{array}{cc}
  \cos{\psi}  & \sin{\psi}\\
  -\frac{1}{l_{r}} \sin{\psi} & \frac{1}{l_{r}} \cos{\psi} \\
  \end{array}
  \right]
    \left[
  \begin{array}{c}
   \dot{x} \\
   \dot{y}  \\
  \end{array}
  \right]
\end{eqnarray}
and the kinematic control law proposed in \cite{Martins20081354} to be applied to the tractor for trajectory tracking control is written as:
\begin{eqnarray}\label{inversekinematicmodeltractor2}
\left[
  \begin{array}{c}
   u_{ref} \\
  \gamma_{ref} \\
  \end{array}
  \right]
  =
  \left[
  \begin{array}{cc}
  \cos{\psi}  & \sin{\psi}\\
  -\frac{1}{l_{r}} \sin{\psi} & \frac{1}{l_{r}} \cos{\psi} \\
  \end{array}
  \right]
    \left[
  \begin{array}{c}
   \dot{x_d} + k_s \tanh{(k_e e_x)}\\
   \dot{y_d} + k_s \tanh{(k_e e_y)} \\
  \end{array}
  \right]
\end{eqnarray}
where $e_x=x_d-x$ and $e_y=y_d-y$ are the current position errors in the axes $X$ and $Y$, respectively. The parameter $k_{e}$ is the gain of the controller and $k_s$ is the saturation constant. The coordinates $(x,y)$ and $(x_d,y_d)$ are the current and the desired coordinates, respectively. The parameters $u_{ref}$ and $\gamma_{ref}$ are the generated references for the speed and the yaw rate controllers.

\subsection{Dynamic Controllers}
\begin{figure*}[h!]
\renewcommand{\arraystretch}{1.1}
\centering
  \includegraphics[width=6in]{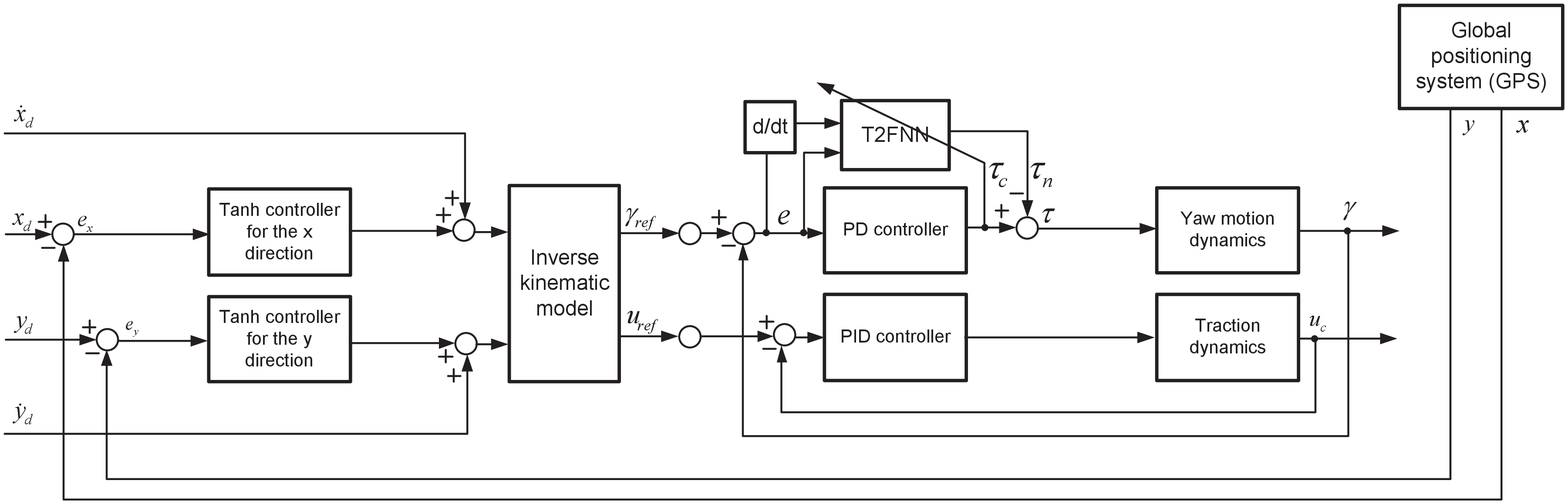}\\
  \caption{Block diagram of the proposed adaptive fuzzy neuro scheme}
  \label{FNN_general}
\end{figure*}

The proposed control scheme used in this study is illustrated in Fig. \ref{FNN_general}. The arrow  in Fig. \ref{FNN_general} indicates that the output of the PD controller is used to tune the parameters of the T2FNN. The output of the PD+T2FNN controller is the steering angle of the front wheel. A low level controller is used to control the steering mechanism. A proportional-integral-derivative (PID) controller is used for the control of longitudinal velocity. For the control of  yaw dynamics, a conventional controller (such as a PD controller) works in parallel with an intelligent controller. One of the main concerns of this study is to implement a novel learning algorithm for T2FNNs by using some novel SMC theory-based learning rules to a real time system. As a testing environment, the yaw dynamics of the tractor in our laboratory has been chosen. On the other hand, the longitudinal dynamics could, of course, be selected.  Moreover, there could be two T2FNNs running on the control of the two subsystems (yaw dynamics and longitudinal dynamics) simultaneously. The reason for such a selection in this study is that the yaw dynamics control of the tractor is more common in agricultural machines. Even if a time-based trajectory (both the yaw angle and the longitudinal speed of the tractor are controlled simultaneously) is given to the system in this study, a space-based trajectory (the longitudinal speed is fixed and only the yaw dynamics is controlled) is also very common in agricultural applications. Based-on these concerns, the yaw dynamics of the tractor has been chosen for the implementation for the novel learning algorithm proposed in this study.

\section{Type-2 Fuzzy-Neuro Control Approach}
\subsection{Type-2 Fuzzy Triangular Membership Functions}
Among the existing MFs in literature (that include a novel one proposed by the authors \cite{ellipsoidal}), triangular MFs are preferred in this paper. Their advantages over the others are mentioned in \cite{Erdal_Triangular}.

\begin{figure}[h!]
\centering
\subfigure[ ]{
\includegraphics[scale=0.25]{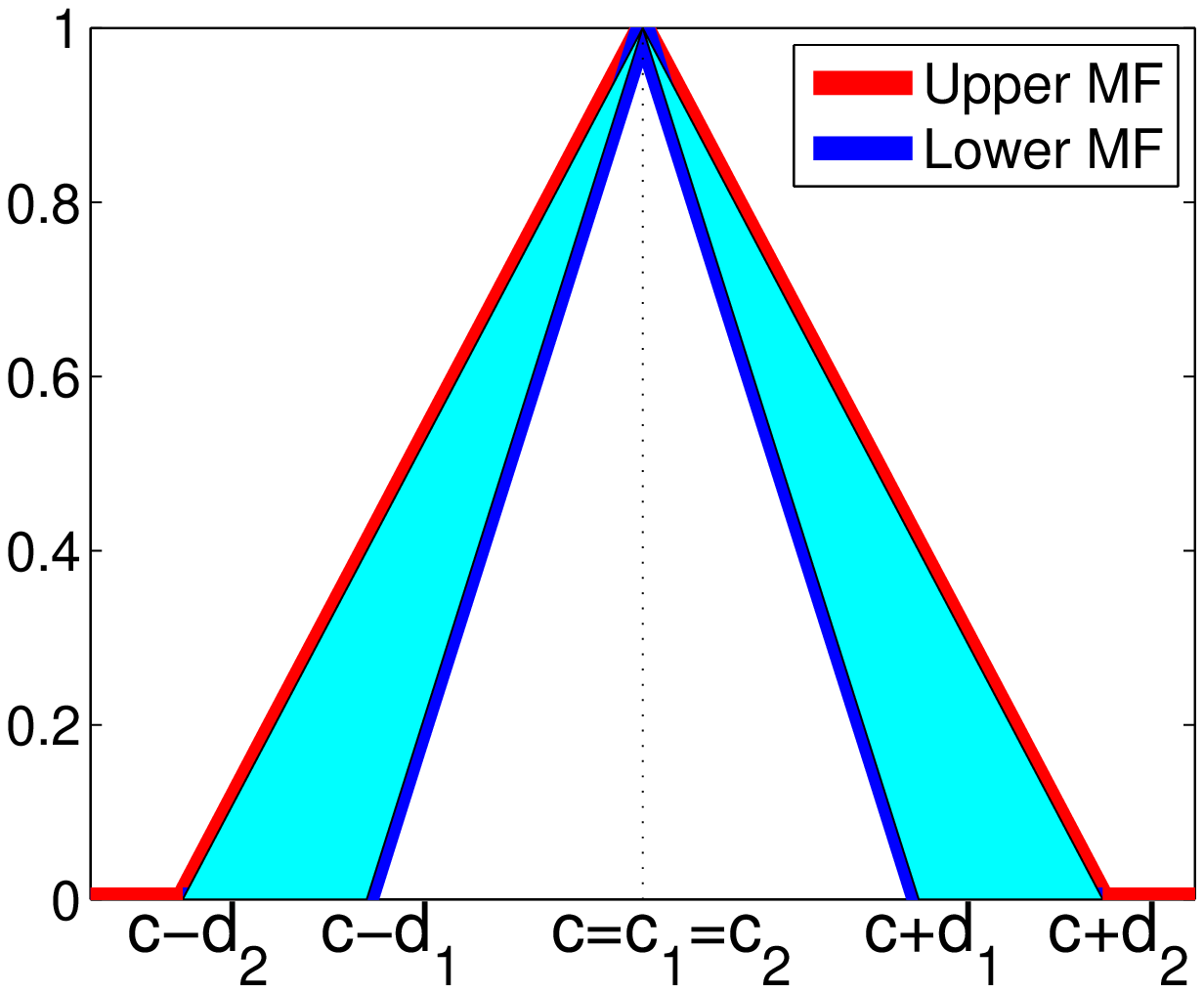}
\label{m1}
}
\subfigure[ ]{
\includegraphics[scale=0.25]{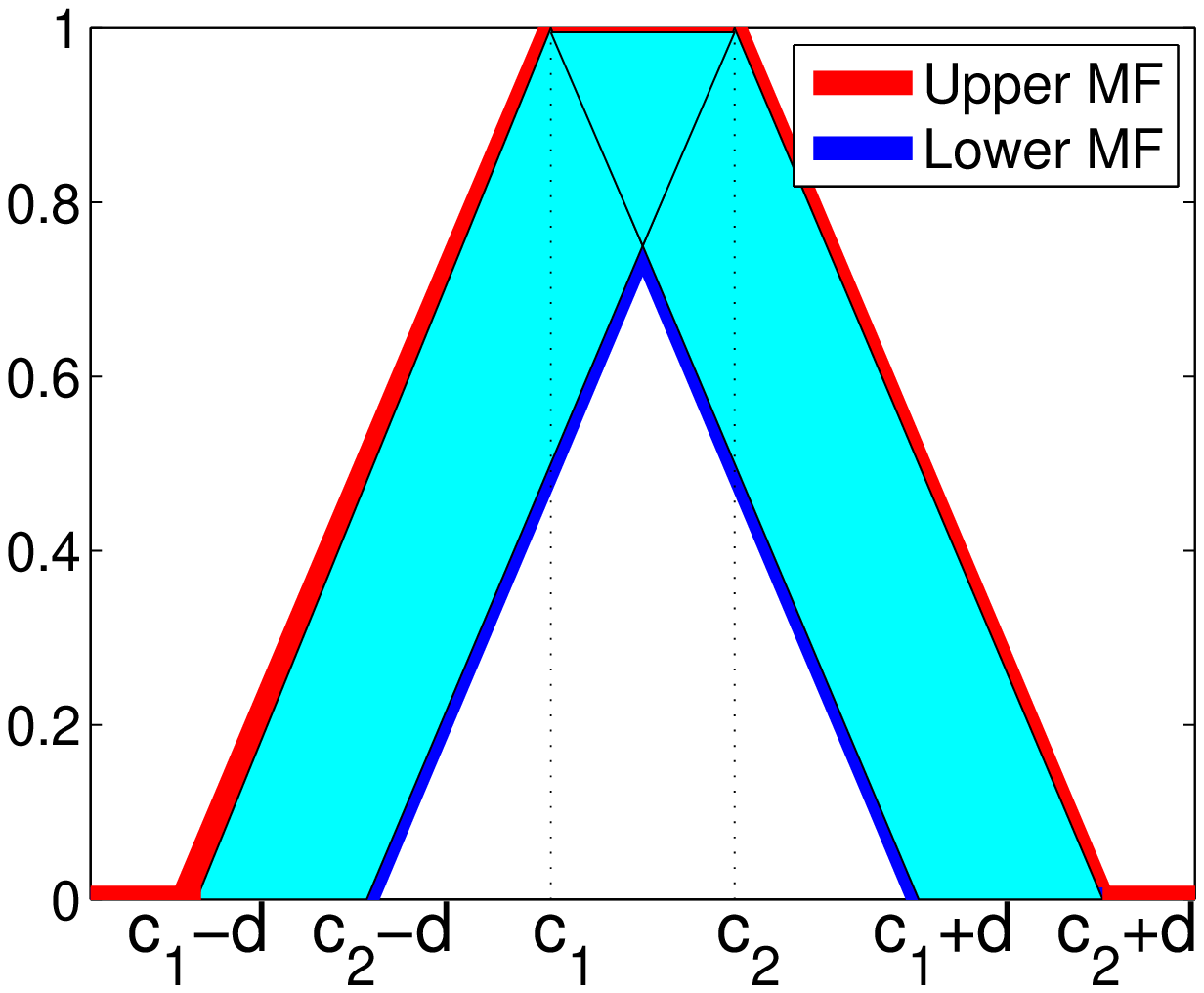}
\label{m2}
}
\caption[Optional caption for list of figures]{A type-2 fuzzy triangular MF with uncertain width (a) and uncertain center (b).}
\label{triangular}
\end{figure}

The mathematical expression for a type-1 fuzzy triangular MF can be written as:
\begin{equation}\label{triangular1}
  {{\mu}({{x}})} = \left\{
  \begin{array}{l l}
1-\left| \frac{{{x}}-{{c}}}{{{d}}} \right| & \left| {{x}}-{{c}} \right|<{{d}}   \\
   0 & otherwise  \\
  \end{array} \right.
\end{equation}
where $c$ and $d$ are the \textit{center} and the \textit{width} of the MF and $x$ is the input vector. On the other hand, type-2 fuzzy triangular MFs with uncertain width and uncertain center are shown in Figs. \ref{m1} and \ref{m2}, respectively. In this paper, MFs with uncertain width are preferred in the antecedent parts of the fuzzy \emph{if-then} rules.

Each MF in the antecedent part is represented by an \emph{upper} (red line in Fig. \ref{triangular}) and a \emph{lower} (blue line in Fig. \ref{triangular}) MF. The membership values for the upper and the lower MFs are denoted as $\overline{\mu }(x){\rm \; and\; \; }\underline{\mu }(x){\rm},$ respectively. The strength of the rule $R_{ij}$ is calculated as a \emph{T}-norm of the MFs in the premise part by using a multiplication operator:
\begin{equation} \label{Rahib2_5}
\underline{W_{ij}}=\underline{\mu_{1i} (x_{1} )} \; \underline{\mu_{2j} (x_{2} )} \;\; \text{and} \;\;
\overline{W_{ij}}=\overline{\mu_{1i} (x_{1} )} \; \overline{\mu_{2j} (x_{2} )}
\end{equation}

The type-2 fuzzy triangular membership values $\underline{\mu_{1i}(x_1)}$, $\overline{\mu_{1i}(x_1)} $, $\underline{\mu_{2j}(x_2)}$, and $\overline{\mu_{2j}(x_2)} $ of the inputs $x_1$ and $x_2$ in the above expression have the
following appearance:

\begin{equation}\label{tri1i_alt}
  \underline{{{\mu }_{1i}}({{x}_{1}})} = \left\{
  \begin{array}{l l}
1-\left| \frac{{{x}_{1}}-{{{c}_{1i}}}}{\underline{{{d}_{1i}}}} \right| & \left| {{x}_{1}}-{{{c}_{1i}}} \right|<\underline{{{d}_{1i}}}  \\
   0 & otherwise  \\
  \end{array} \right.
\end{equation}

\begin{equation*}
  \overline{{{\mu }_{1i}}({{x}_{1}})} = \left\{
  \begin{array}{l l}
1-\left| \frac{{{x}_{1}}-{{{c}_{1i}}}}{\overline{{{d}_{1i}}}} \right| & \left| {{x}_{1}}-{{{c}_{1i}}} \right|<\overline{{{d}_{1i}}}   \\
   0 & otherwise  \\
  \end{array} \right.
\end{equation*}

\begin{equation*}
  \underline{{{\mu }_{2j}}({{x}_{2}})} = \left\{
  \begin{array}{l l}
1-\left| \frac{{{x}_{2}}-{{{c}_{2j}}}}{\underline{{{d}_{2j}}}} \right| & \left| {{x}_{2}}-{{{c}_{2j}}} \right|<\underline{{{d}_{2j}}}  \\
   0 & otherwise  \\
  \end{array} \right.
\end{equation*}

\begin{equation*}
  \overline{{{\mu }_{2j}}({{x}_{2}})} = \left\{
  \begin{array}{l l}
1-\left| \frac{{{x}_{2}}-{{{c}_{2j}}}}{\overline{{{d}_{2j}}}} \right| & \left| {{x}_{2}}-{{{c}_{2j}}} \right|<\overline{{{d}_{2j}}}  \\
   0 & otherwise  \\
  \end{array} \right.
\end{equation*}

\subsection{Interval Type-2 A2-CO TSK Model}
\begin{figure}[h!]
\centering
  \includegraphics[width=3.5in]{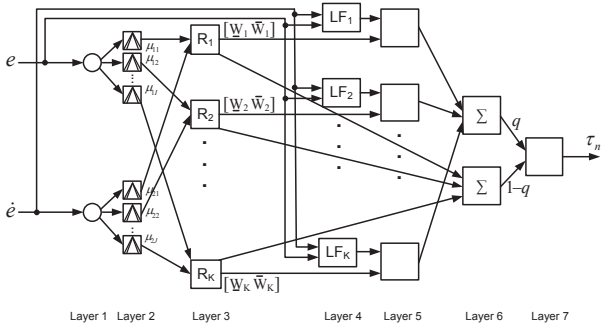}\\
  \caption{The structure of the proposed T2FNN for two inputs}
  \label{fig_T2FNN}
\end{figure}

The interval T2FLS considered in this paper uses type-2 triangular MFs in the premise part and crisp numbers in the consequent part. This structure is called A2-C0 fuzzy system \cite{Biglarbegian}, and it is shown in Fig. \ref{fig_T2FNN}. The fuzzy \emph{If-Then} rule $R_{ij}$ of a $zero^{th}$-order type-2 TSK model with two input variables in which the consequent part is a crisp number can be defined as follows:

\begin{equation}
R_{ij}: \;\; \text{If} \; x_1 \; \text{is} \;\; \tilde{A}_{1i} \;\; \text{and} \; x_2 \; \text{is} \;\; \tilde{A}_{2j}, \;\; \text{then} \; f_{ij}=d_{ij}
\end{equation}

The followings are the operations in each Layer in Fig. \ref{fig_T2FNN}: In Layer 1, the input signals feed the system. The related figure shows the system for two inputs which are the error and the time derivative of the error. In Layer 2, the membership degrees $\underline{\mu }$ and $\overline{\mu }$ are determined for each input signal entering the system. Layer 3 calculates the firing strengths of the rules which are realized using  the \emph{prod} t-norm operator using \eqref{Rahib2_5}. Layer 4 determines the outputs of the linear functions $f_{ij}$ $(i=1,\dots,I$ and $j=1,\dots,J)$, in the consequent parts for the two inputs case.

\begin{equation} \label{Rahib2_7}
f_{ij} = d_{ij}
\end{equation}

Layer 5 computes the product of the membership degrees $\underline{W_{ij}}{\rm \; \; and\; \; }\overline{W_{ij}}{\rm \; }$ and linear functions $f_{ij}$. Two summation blocks are in Layer 6. One of these blocks computes the sum of the output signals from Layer 5 (the numerator part of \eqref{4}) and the other block computes the sum of the output signal of Layer 3 (the denominator part of \eqref{4}). Finally, Layer 7 calculates the output of the network using \eqref{Rahib2_6}.

The output of the network is calculated as follows:
\begin{equation}\label{4}
\tau_n=\int_{W_{11}\epsilon[\underline{W}_{11},\overline{W}_{11}]}\ldots\int_{W_{IJ}\epsilon[\underline{W}_{IJ},\overline{W}_{IJ}]}1\bigg/\frac{\sum_{i=1}^{I}\sum_{j=1}^{J}W_{ij}(x)f_{ij}}{\sum_{i=1}^{I}\sum_{j=1}^{J}W_{ij}(x)}
\end{equation}
where $f_{ij}$ is given by the \emph{If-Then} rule. The inference engine used in this paper replaces the type-reduction which is given as:

\begin{equation}\label{Rahib2_6}
\tau_n=\frac{q(t)\sum _{i=1}^{I}\sum _{j=1}^{J}\underline{W}_{ij} f_{ij}  }{\sum _{i=1}^{I}\sum _{j=1}^{J}\underline{W}_{ij}  } +\frac{\big(1-q(t)\big)\sum _{i=1}^{I}\sum _{j=1}^{J}\overline{W}_{ij} f_{ij}  }{\sum _{i=1}^{I}\sum _{j=1}^{J}\overline{W}_{ij}  }
\end{equation}

The design parameter $q$, weights the sharing of the lower and the upper firing levels of each fired rule \cite{Biglarbegian}.  This parameter can be a constant (equal to $0.5$ in most cases) or a time varying parameter. In this paper, the latter is preferred. In other words, the parameter update rules and the proof of the stability of the learning process are given for the case of a time varying $q$.

After the normalization of (\ref{Rahib2_6}), the output signal of the T2FNN will obtain the following form:

\begin{equation}\label{eq19}
\tau_n=q(t) \sum_{i=1}^{I}\sum_{j=1}^{J}f_{ij}\widetilde{\underline{W_{ij}}}+\big(1-q(t)\big)\sum_{i=1}^{I}\sum_{j=1}^{J}f_{ij}\widetilde{\overline{W_{ij}}}
 \end{equation}

\noindent where $\widetilde{\underline{W_{ij}}}$ and $\widetilde{\overline{W_{ij}}}$ are the normalized values of the lower and the upper output signals of the neuron $ij$ from the second hidden layer of the network:
\begin{eqnarray}
\widetilde{\underline{{W_{ij}}}} = \frac{\underline{W_{ij}}}{\sum_{i=1}^{I}\sum_{j=1}^{J}\underline{W_{ij}}} \mbox{ and }
\widetilde{\overline{{W_{ij}}}} = \frac{\overline{W_{ij}}}{\sum_{i=1}^{I}\sum_{j=1}^{J}\overline{W_{ij}}} \nonumber
\end{eqnarray}

The following vectors can be specified:

\noindent $\widetilde{\underline{{W}}}\left(t\right)=\left[\widetilde{\underline{{W_{11}}}} \left(t\right)\; \widetilde{\underline{{W_{12}}}} \left(t\right)...\; \widetilde{\underline{{W_{21}}}}
\left(t\right)\; ...\;\widetilde{\underline{{W_{ij}}}} \left(t\right)\; ...\;\widetilde{\underline{W}_{IJ}} \left(t\right)\right]^{T} $

\noindent $\widetilde{\overline{{W}}}\left(t\right)=\left[\widetilde{\overline{{W_{11}}}} \left(t\right)\; \widetilde{\overline{{W_{12}}}} \left(t\right)...\; \widetilde{\overline{{W_{21}}}}
\left(t\right)\; ...\;\widetilde{\overline{{W_{ij}}}} \left(t\right)\; ...\;\widetilde{\overline{W}_{IJ}} \left(t\right)\right]^{T} $

\noindent $F=[f_{11} \; f_{12} \; ...\; f_{21} \; ...\; f_{ij}\; ...\; f_{IJ}] $

The following assumptions have been used in this investigation: Both the input signals $x_{1}(t)$ and $x_{2}(t)$, and their time derivatives can be considered bounded:

\begin{equation}
|x_1(t)|\leq \widetilde{B_x}, \hspace{0.25cm} |x_2(t)|\leq \widetilde{B_x}   \hspace{0.5cm}\forall t
\end{equation}

\begin{equation}
|\dot{x}_1(t)|\leq \widetilde{B_{\dot{x}}}, \hspace{0.25cm} |\dot{x}_2(t)|\leq \widetilde{B_{\dot{x}}} \hspace{0.5cm}\forall t
\end{equation}
where $\widetilde{B_x}$ and $\widetilde{B_{\dot{x}}}$ are assumed to be some known positive constants. It is obvious that $0<\widetilde{\underline{{W_{ij}}}}\leq1$ and $0<\widetilde{\overline{{W_{ij}}}}\leq1$. In addition, it can be easily seen that $\sum_{i=1}^{I}\sum_{j=1}^{J}\widetilde{\underline{{W_{ij}}}}=1$ and $\sum _{i=1}^{I}\sum_{j=1}^{J}\widetilde{\overline{{W_{ij}}}}=1$. It is also considered that, $\tau$ and $\dot{\tau}$ will be bounded signals too, \emph{i.e.}
\begin{equation}
\left|\tau(t)\right|< B_{\tau},\hspace{0.25cm}  \left|\dot{\tau}\left(t\right)\right|< B_{\dot{\tau }} \quad \forall t
\end{equation}
where $B_{\tau} $ and $B_{\dot{\tau}} $ are some known positive constants.

\subsection{The SMC Theory-based Learning Algorithm}
Using the principles of SMC theory \cite{Utkin} the zero value of the learning error coordinate $\tau _{c} \left(t\right)$ can be
defined as a time-varying sliding surface, \emph{i.e.},

\begin{equation}
S_{c} \left(\tau _{n} ,\tau \right)=\tau _{c} \left(t\right)=\tau _{n} \left(t\right)+\tau \left(t\right)=0
\end{equation}
which is the condition that the T2FNN is trained to become a nonlinear regulator to obtain the desired response during the tracking-error convergence movement by compensating the nonlinearity of the controlled plant.

The sliding surface for the nonlinear system under control $S_{p} \left(e,\dot{e}\right)$ is defined as:

\begin{equation}
S_{p} \left(e,\dot{e}\right)=\dot{e}+\chi e
\end{equation}
with $\chi$ being a positive constant determining the slope of the sliding surface.

\textit{Definition:} A sliding motion will appear on the sliding manifold $S_{c}\left(\tau_{n},\tau\right)=\tau_{c}\left(t\right)=0$
after a time $t_{h} $, if the condition $S_{c} (t)\dot{S}_{c} (t)=\tau _{c} \left(t\right)\dot{\tau }_{c} \left(t\right)<0$ is satisfied for all
$t$ in some nontrivial semi-open subinterval of time of the form $\left[t,t_{h} \right)\subset \left(0, t_{h} \right)$.

It is desired to devise a dynamical feedback adaptation mechanism, or an online learning algorithm for the parameters of the T2FNN considered, such that the sliding mode condition of the above definition is enforced.

\subsection{The Proposed Parameter Update Rules for the T2FNN}
The parameter update rules for the T2FNN proposed in this paper which has two inputs are given by the following theorem.

\emph{Theorem 1:} If the adaptation laws for the parameters of the considered T2FNN are chosen as:

\begin{equation}\label{16}
\underline{\dot c_{1i}} = \overline{\dot c_{1i}}={\dot c_{1i}}=\dot x_1
\end{equation}
\begin{equation}\label{16ek}
\underline{\dot c_{2j}} = \overline{\dot c_{2j}}={\dot c_{2j}}=\dot x_2
\end{equation}
\begin{equation}\label{enyeni1}
\underline{\dot d_{1i}}=\underline{\mu_{1i}}\frac{-\alpha \underline{d_{1i}}^2}{x_1-{c_{1i}}} sgn(\tau_c) sgn\bigg(\frac{x_1-{c_{1i}}}{\underline{d_{1i}}}\bigg)
\end{equation}
\begin{equation}
\overline{\dot d_{1i}}=\overline{\mu_{1i}}\frac{-\alpha \overline{d_{1i}}^2}{x_1-{c_{1i}}} sgn(\tau_c) sgn\bigg(\frac{x_1-{c_{1i}}}{\overline{d_{1i}}}\bigg)
\end{equation}
\begin{equation}
\underline{\dot d_{2j}}=\underline{\mu_{2j}}\frac{-\alpha \underline{d_{2j}}^2}{x_2-{c_{2j}}} sgn(\tau_c) sgn\bigg(\frac{x_2-{c_{2j}}}{\underline{d_{2j}}}\bigg)
\end{equation}
\begin{equation}
\overline{\dot d_{2j}}=\overline{\mu_{2j}}\frac{-\alpha \overline{d_{2j}}^2}{x_2-{c_{2j}}} sgn(\tau_c) sgn\bigg(\frac{x_2-{c_{2j}}}{\overline{d_{2j}}}\bigg)
\end{equation}

\begin{equation}\label{fij}
\dot{f_{ij}}= -\frac{\Big(q(t)\underline{\widetilde{W_{ij}}}+ \big(1-q(t)\big)\widetilde{\overline{W_{ij}}}\Big)\alpha sgn(\tau_c)}{\Big(q(t)\underline{\widetilde{W}}+\big(1-q(t)\big) \widetilde{\overline{W}}\Big)^T\Big(q(t)\underline{\widetilde{W}}+ \big(1-q(t)\big)\widetilde{\overline{W}}\Big)}
\end{equation}
\begin{equation}\label{q_q}
\dot{q}(t)= -\frac{\alpha sgn(\tau_c)}{F(\underline{\widetilde{W}}-\widetilde{\overline{W}})^T}
\end{equation}
where $\alpha$ is a sufficiently large positive design constant satisfying the inequality below:
\begin{equation}\label{alphaa}
\alpha >B_{\dot{\tau }}
\end{equation}
Then, given an arbitrary initial condition $\tau_c(0)$, the learning error $\tau_c(t)$ will converge  to zero within a finite time $t_h$.

\textit{Proof: }The reader is referred to  Appendix A.

In order to avoid division by zero in the adaptation laws of \eqref{enyeni1}-\eqref{q_q} an instruction is included in the algorithm to make the denominator equal to $0.001$ when its calculated value is smaller than this threshold.

It is well-known that sliding mode control suffers from high-frequency oscillations in the control input, which are called \emph{chattering}. The following are the two common methods used to eliminate chattering \cite{Slotine}:
\begin{enumerate}
  \item Using a saturation function to replace the signum function.
  \item Inserting a boundary layer so that an equivalent control replaces the corrective one when the system is inside this layer.
\end{enumerate}

In order to reduce the chattering effect, the following function is used in this paper with $\delta_s=0.05$ instead of the signum function in the dynamic strategy described in \eqref{enyeni1}-\eqref{q_q}.
\begin{equation}\label{chatter}
\textrm{sgn}(\tau_{c}):=\frac{\tau_{c}}{|\tau_{c}|+\delta_s}
\end{equation}
where $\delta_s=0.05$.

The relation between the sliding line $S_{p}$ and the zero adaptive learning error level $S_{c}$ is determined by the following equation:
\begin{equation}
S_{c} =\tau _{c} =k_{D} \dot{e}+k_{P} e=k_{D} \left(\dot{e}+\frac{k_{p} }{k_{D} } e\right)=k_{D} S_{p}
\end{equation}

The tracking performance of the feedback control system can be analyzed by introducing the following Lyapunov function candidate:
\begin{equation}\label{lyapunov}
V_{p} =\frac{1}{2} S_{p}^{2}
\end{equation}

\textit{Theorem 2}: If the adaptation strategy for the adjustable parameters of the T2FNN is chosen as in (\ref{16})-(\ref{q_q}), then the
negative definiteness of the time derivative of the Lyapunov function in (\ref{lyapunov}) is ensured.

\textit{Proof: }The reader is referred to  Appendix B.

\emph{Remark}: The obtained result means that, assuming that the SMC task is achievable, using $\tau_{c}$  as a learning error for the T2FNN together with the adaptation laws (\ref{16})-(\ref{q_q}) enforces the desired reaching mode followed by a sliding regime for the system under control. 
\section{Experimental Setup}
\begin{figure}[h!]
\centering
\includegraphics[width=3.5in]{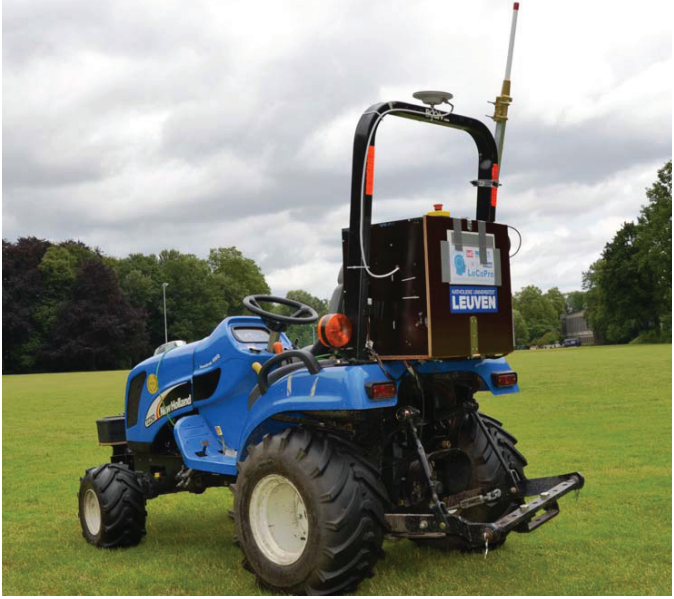}\\
\caption{The autonomous tractor}
\vskip -0.1cm
\label{tractor1}
\end{figure}

The global objective in the following real-time experiments is to track a time-based trajectory with the Case New Holland TZ25DA tractor shown in Fig. \ref{tractor1}.

\subsection{Global Positioning System}
A global positioning system (GPS) antenna is located straight up the center of the tractor rear axle to provide highly accurate positional information for the autonomous tractor. The height of the antenna is 2m above ground level. It is connected to a Septentrio AsteRx2eH RTK-DGPS receiver (Septentrio Satellite Navigation NV, Belgium) with a specified position accuracy of 2cm at a 20-Hz sampling frequency (sampling period $T_{s}=0.05$ s). The RTK correction signals are obtained from the Flepos network through a wireless internet connection established with a Digi Connect WAN 3G modem.

\subsection{Hardware, Software and Sensors}
\begin{figure}[h!]
\centering
\includegraphics[width=3.5in]{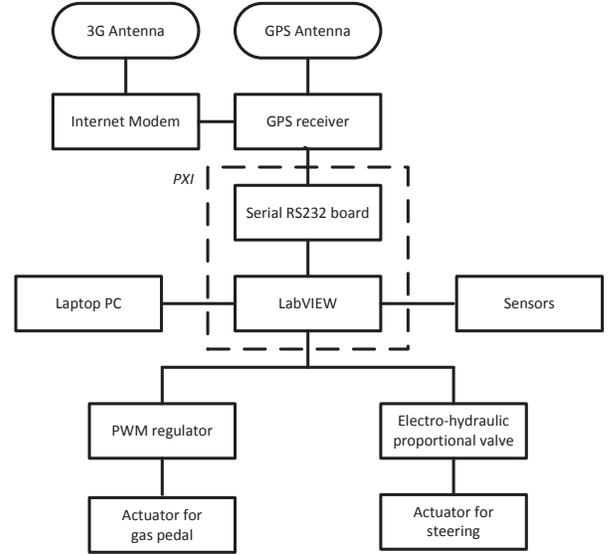}\\
\caption{Block diagram of the hardware}
\vskip -0.1cm
\label{blockdiagramhardware2}
\end{figure}

The block diagram of hardware is shown in Fig. \ref{blockdiagramhardware2}. The GPS receiver and the internet modem are connected to a real time operating system (PXI platform, National Instrument Corporation, USA) via a RS232 serial communication. The PXI system gathers the steering angles and the GPS data, and it controls the tractor by sending signals to the actuators. A laptop connected via a wireless network to the PXI system functions as the user interface of the autonomous tractor. The control algorithms are implemented in $LabVIEW^{TM}$ version 2011, National Instrument, USA. They are executed in real time on the PXI and updated at a rate of 20-Hz.

The PD+T2FNN control algorithm calculates the desired steering angle for the front wheels. In the inner closed loop, the steering mechanism is controlled by using an electro-hydraulic valve from Sauer Danfoss with a flow of 12 liter/min. The electro-hydraulic valve characteristics are highly nonlinear and include a saturation and a dead-band region. The voltage, limited between $0-12$ volt and the steering angle, limited between $\pm45^{\circ}$ constitute the input and the output for the steering system, respectively. The position of the front wheels is measured using a potentiometer mounted on the front axle, yielding a position measurement with a resolution of $1^{\circ}$. The position of the front wheel is measured 100 times at every 50 ms. The data are filtered by using a Savitzky-Golay filter which tends to preserve features of the distribution such as relative maxima, minima and width. After filtering process, the steering angle is found by taking the mean value of the filtered data. This procedure is repeated every 50 ms.

The speed of the tractor is controlled by using an electro-mechanic valve. Two PID type controllers are used in cascade fashion. The one in the outer loop generations the desired pedal position with respect to the speed of the tractor and the one in the inner loop generates the voltage value for the electro-mechanic valve with respect to the pedal position. Like in the measuring of the steering angle, the pedal position is measured 100 times at every 50 ms and the data are filtered by using a Savitzky-Golay filter and taking the mean value of the filtered data. Since the measured speed signal coming from the GPS is noisy, a discrete Kalman filter (KF) is used to reduce noise. A position-velocity model described in \cite{Brown} is used where vehicle velocity is assumed as a random-walk process. The KF assumes that the vehicle moves with a constant velocity between discrete-time steps. The state vector of the model used in the KF and the state transition matrix are as follows:
\begin{eqnarray}
\label{kinematicmodelKF}
\widehat{x}_{k+1} & = & \Phi(T_{s}) \widehat{x}_{k} \nonumber \\
  & = & \left[
  \begin{array}{cccc}
   1 & T_{s} & 0 & 0 \\
   0 & 1 & 0 & 0 \\
   0 & 0 & 1 & T_{s}  \\
   0 & 0 & 0 & 1  \\
  \end{array}
  \right]
   \left[
  \begin{array}{c}
   x_{k}  \\
   v_{x,k}  \\
   y_{k}  \\
   v_{y,k}  \\
  \end{array}
  \right]
\end{eqnarray}
where $\Phi(T_{s})$, $v_{x,k}$ and $v_{y,k}$ are the state transition matrix, easting and northing velocities coming from the GPS, respectively.

\subsection{State Estimation}
Some states of the autonomous tractor cannot be measured. Even when the states can be measured directly, the measurements will contain delay and noise. Moreover, at some time steps, no useful position data are obtained from the GPS receiver.

An extended Kalman filter (EKF) is used for state estimation. Since the GPS antenna is located at the R point on the tractor, the kinematic model in (\ref{kinematicmodeltractor}) is used. The discrete-time kinematic model used by the EKF is written with a sampling interval $T_{s}$ as follows:
\begin{eqnarray}
\label{kinematicmodelEKF}
x_{k+1} & = & x_{k} + T_{s} u_{k} \cos{\psi_{k}} \nonumber \\
y_{k+1} & = & y_{k} + T_{s} u_{k} \sin{\psi_{k}} \nonumber \\
\psi_{k+1} & = & \psi_{k} + T_{s} u_{k} \frac{\tan{\delta_{k}}}{L}
\end{eqnarray}

The general form of the estimated system model is:
\begin{eqnarray}
\label{generalmodelEKF}
\widehat{x}_{k+1} & = & f(\widehat{x}_{k}, u_{k}) + w_{k} \nonumber \\
\widehat{y}_{k+1} & = & h(\widehat{x}_{k}) + v_{k}
\end{eqnarray}
where $f$ is the estimation model for the system and $h$ is the measurement function. The differences between the kinematic model and the estimation model are the process noise $w_{k}$ and the observation noise $v_{k}$ both in the state and the measurement equations. They are both assumed to be independent with zero mean multivariate Gaussian noises with covariance matrices $Q_{k}$ and $R_{k}$, respectively:
\begin{eqnarray}
\label{noise}
w_{k} \backsim N(0,Q_{k}) \nonumber \\
v_{k} \backsim N(0,R_{k})
\end{eqnarray}

Since only one GPS antenna was mounted on the tractor, the yaw angle of the tractor was not measured directly. It is to be noted that the knowledge of the yaw angle of the tractor plays a very important role in the accuracy of trajectory tracking control. The estimated value of the yaw angle is used in the inverse kinematic model to generate the desired speed and the desired yaw rate for the system. The inputs of the EKF are the position and the velocity, the velocity values from GPS and the steering angle value from the potentiometer in the front wheels. The outputs of the EKF are the position of the tractor in the x- and the y-coordinates system and the yaw angle. These estimated values are then used in the trajectory control.

\subsection{Experimental Results}
\begin{figure}[h!]
\centering
\includegraphics[width=3.5in]{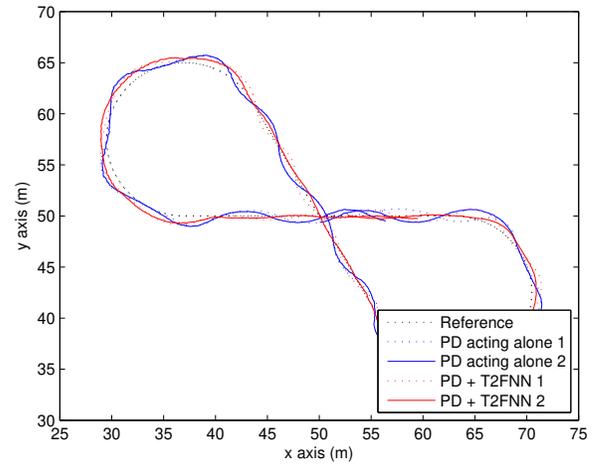}\\
\caption{The trajectory of the autonomous tractor}
\vskip -0.1cm
\label{trajj}
\end{figure}

\begin{figure}[h!]
\centering
\includegraphics[width=3.5in]{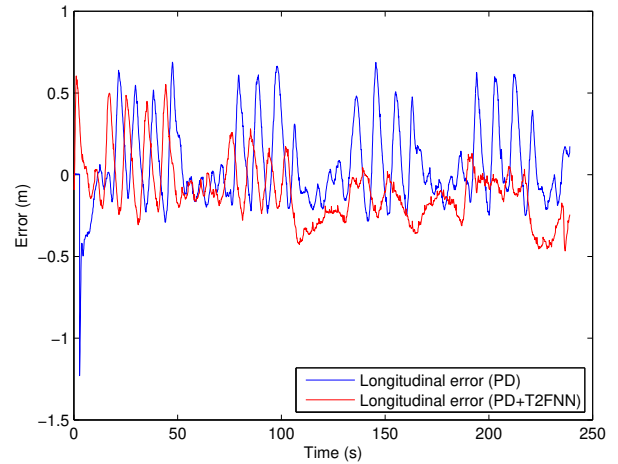}\\
\caption{The longitudinal error}
\vskip -0.1cm
\label{error_long}
\end{figure}

\begin{figure}[h!]
\centering
\includegraphics[width=3.5in]{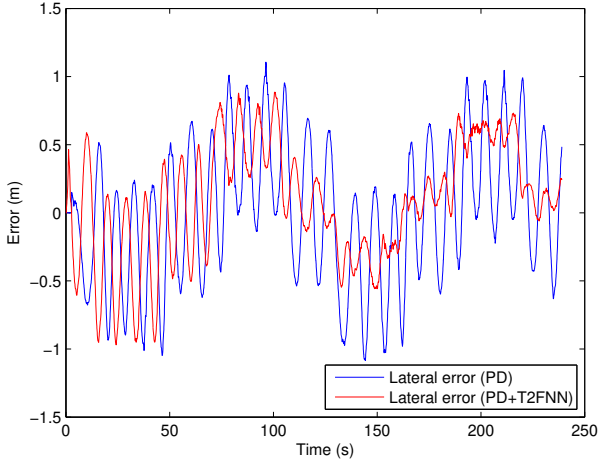}\\
\caption{The lateral error}
\vskip -0.1cm
\label{error_lat}
\end{figure}

An 8-shaped reference trajectory is applied to the system. The reference and the actual trajectories of the system, both the longitudinal and the lateral error values on the related trajectory are shown in Fig. \ref{trajj}, Fig. \ref{error_long} and Fig. \ref{error_lat} for two different controllers, respectively. The results show that the control scheme consisting of a T2FNN working in parallel with a PD controller gives a better trajectory following accuracy than the one where only a PD controller acts alone. It can here be argued that the performance of  the conventional controller acting alone can be improved by better tuning, but as it has already been stated that in real life, this is a challenging task; because in addition to the interactions of the subsystems, there exist unmodeled dynamics and uncertainties in real world applications. Thus, the proposed control structure, consisting of an intelligent controller and a conventional controller would be preferable in real life.

In Fig. \ref{trajj}, while the dotted lines show the first turns, the solid lines represent the second turns. The control accuracy for the PD controller is the same for the first and second turn which is the expected case. However, the control accuracy for the T2FNN working in parallel with a PD controller is better for the second turn. We can observe similar behaviour in Fig. \ref{error_long} and Fig. \ref{error_lat}. The mean square of the lateral error in Fig. \ref{error_lat} is equal to 0.2575 and 0.1803 for the case of the PD controller working alone and the case of the PD controller working in parallel with the T2FNN, respectively. The results show a performance improvement of 30\% in the latter case.

\begin{figure}[h!]
\centering
\includegraphics[width=3.5in]{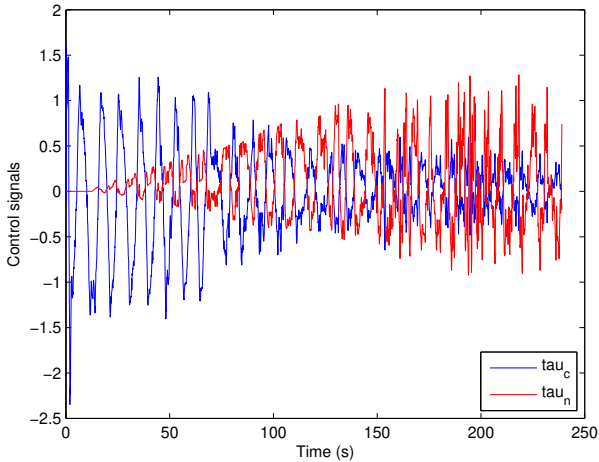}\\
\caption{The control signals}
\vskip -0.1cm
\label{controlsig}
\end{figure}

Figure \ref{controlsig} shows the control signals coming from the conventional PD controller and T2FNN when the PD controller works in parallel with the T2FNN. As can be seen from Fig. \ref{controlsig}, at the beginning (in the first turn), the dominating control signal is the one coming from the PD controller. After the first turn (starting from $120^{th}$ second), the T2FNN is able to take over the control, thus becoming the leading controller. Moreover, when the reference signal changes, the output of the PD controller increases. However, after a finite time, the output of the PD controller again comes back to approximately zero.

\begin{figure}[h!]
\centering
\includegraphics[width=3.5in]{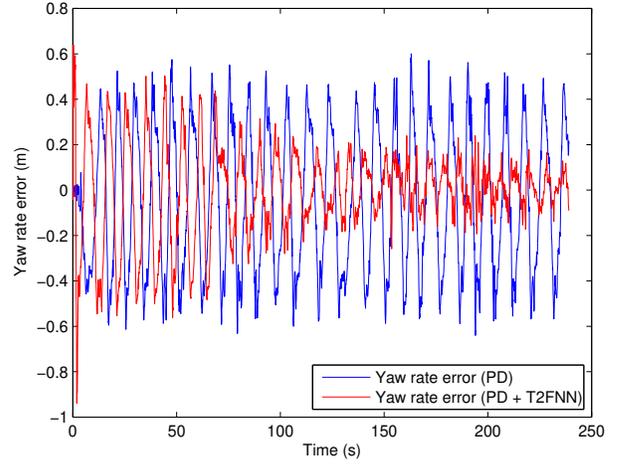}\\
\caption{The yaw rate error}
\vskip -0.1cm
\label{erroryawrate}
\end{figure}

Although there exist two independent subsystem controllers in the autonomous tractor control system, the T2FNN works in parallel with a PD controller only for the control of the yaw rate of the system. Thus, the error signals for both PD controller acting alone and in parallel with the T2FNN are shown in Fig. \ref{erroryawrate}. As can be seen from Fig. \ref{erroryawrate}, the T2FNN significantly increases the control accuracy of the yaw dynamics of the system.

\begin{figure}[h!]
\centering
\includegraphics[width=3.5in]{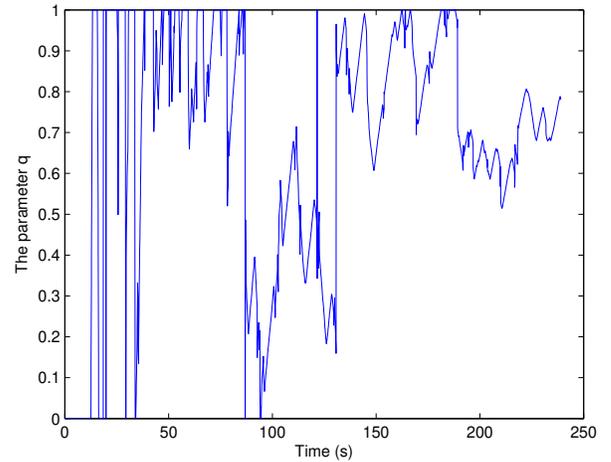}\\
\caption{The parameter q}
\vskip -0.1cm
\label{q}
\end{figure}

Figure \ref{q}  represents the online tuning of the parameter $q$ in \eqref{eq19} which weights the sharing of the lower and the upper firing levels of each fired rule. The learning algorithms derived in this paper includes a parameter adaptation rule for $q$ too and we can see the effect of this on this figure when the reference trajectory changes abruptly. It can also be observed that the change in this parameter is not very big in the second turn when compared to the first turn. During the training of the parameter q, a constraint is put to the parameter in which it is bounded between 0 and 1. When it reaches its limits, the tuning is turned off for that parameter.

\section{Conclusions}
In an interval TSK T2FLS, there exists a design parameter that balances the sharing of the lower and the upper firing levels of each fired rule, and this parameter can be tuned during the real-time control of the system. One novelty of this paper is that an SMC theory-based learning algorithm is proposed for the tuning of that parameter, the effect of which is  evaluated in real-time for the control of the yaw dynamics of an autonomous tractor.  Instead of forcing the designer to choose a fixed value for the above mentioned parameter, this study allows the system to tune it online providing an additional degree of freedom for the overall controller. The real-time results show that when the T2FNN is used in parallel with a conventional PD controller, the overall system learns the system dynamics to perform a better performance in terms of a smaller settling time and near zero steady state error. The use of the combination of fuzzy logic control, artificial neural networks and sliding mode control theory harmoniously allows us to better handle the interactions in the subsystems, uncertainties and lack of modeling information. In addition to its well-known robustness property, another prominent feature of the proposed approach is its computational simplicity.

\section*{Acknowledgment}
This work has been carried out within the framework of projects IWT-SBO 80032 (LeCoPro) of the Institute for the Promotion of Innovation through Science and Technology in Flanders (IWT-Vlaanderen). We would like to thank Mr. Soner Akpinar for his technical support for the preparation of the experimental set up. 
\appendices
\section{Proof of Theorem 1}
The time derivatives of (\ref{tri1i_alt}) are as follows:
\begin{equation}
\underline{\dot \mu _{1i}(x_1)}=\bigg(-\frac{\dot x_1-{\dot c_{1i}}}{\underline{d_{1i}}}+\frac{x_1-{c_{1i}}}{\underline{d_{1i}^2}}\underline{\dot d_{1i}}\bigg) sgn\bigg(\overbrace{\frac{x_1-{c_{1i}}}{\underline{d_{1i}}}}^\text{$\underline{E_1}$}\bigg)
\end{equation}
\begin{equation}
\overline{\dot \mu _{1i}(x_1)}=\bigg(-\frac{\dot x_1-{\dot c_{1i}}}{\overline{d_{1i}}}+\frac{x_1-{c_{1i}}}{\overline{d_{1i}^2}}\overline{\dot d_{1i}}\bigg) sgn\bigg(\overbrace{\frac{x_1-{c_{1i}}}{\overline{d_{1i}}}}^\text{$\overline{E_1}$}\bigg)
\end{equation}
\begin{equation}
\underline{\dot \mu _{2j}(x_2)}=\bigg(-\frac{\dot x_2-{\dot c_{2j}}}{\underline{d_{2j}}}+\frac{x_2-{c_{2j}}}{\underline{d_{2j}^2}}\underline{\dot d_{2j}}\bigg) sgn\bigg(\overbrace{\frac{x_2-{c_{2j}}}{\underline{d_{2j}}}}^\text{$\underline{E_2}$}\bigg)
\end{equation}
\begin{equation}
\overline{\dot \mu _{2j}(x_2)}=\bigg(-\frac{\dot x_2-{\dot c_{2j}}}{\overline{d_{2j}}}+\frac{x_2-{c_{2j}}}{\overline{d_{2j}^2}}\overline{\dot d_{2j}}\bigg) sgn\bigg(\overbrace{\frac{x_2-{c_{2j}}}{\overline{d_{2j}}}}^\text{$\overline{E_2}$}\bigg)
\end{equation}

The time derivative of the strength of the rule $R_{ij}$ for the lower MF in (\ref{Rahib2_5}) is as follows:
\begin{eqnarray}\label{kirkyedi}
\lefteqn{\dot{\widetilde{\underline{W_{ij}}}} = \frac{\Big(\overbrace{\underline{\dot{\mu}_{1i}(x_1)} \underline{\mu_{2j}(x_2)}+ \underline{\mu_{1i}(x_1)} \underline{\dot{\mu}_{2j}(x_2)}}^\text{A}\Big)}{\sum_{i=1}^{I}\sum_{j=1}^{J}\underline{W_{ij}}} } \\
&& {}-\frac{\widetilde{\underline{W_{ij}}}\bigg(\sum_{i=1}^{I}\sum_{j=1}^{J} \Big(\overbrace{\underline{\dot{\mu}_{1i}(x_1)} \underline{\mu_{2j}(x_2)}+ \underline{\mu_{1i}(x_1)} \underline{\dot{\mu}_{2j}(x_2)}}^\text{A}\Big)\bigg)}{\sum_{i=1}^{I}\sum_{j=1}^{J}\underline{W_{ij}}} \nonumber
\end{eqnarray}

If the parameter update rules in Theorem 1 are used, (\ref{kirksekiz}) is achieved:

\begin{eqnarray}\label{kirksekiz}
A &=& \bigg(\Big(\frac{x_1-{c_{1i}}}{\underline{d_{1i}^2}}\Big)
\dot{\underline{d_{1i}}}-\cancelto{0}{\Big(\frac{\dot{x_1}-{\dot{c_{1i}}}}{\underline{\dot{d_{1i}}}}}\Big)\bigg)sgn(\underline{E_1})\big(\overbrace{1-\underline{E}_2sgn(\underline{E_2})}
^\text{$\underline{\mu_{2j}}$}\big) \nonumber \\
&+& {}\bigg(\Big(\frac{x_2-{c_{2j}}}{\underline{d_{2j}^2}}\Big)
\dot{\underline{d_{2j}}}-\cancelto{0}{\Big(\frac{\dot{x_2}-{\dot{c_{2j}}}}
{\underline{\dot{d_{2j}}}}}\Big)\bigg)sgn(\underline{E_2})\big(\overbrace{1-\underline{E_1}sgn(\underline{E_1})}^\text{$\underline{\mu_{1i}}$}\big) \nonumber \\
&=& {} -2\alpha sgn(\tau_c)\underline{\mu_{1i}}\underline{\mu_{2j}} \nonumber \\
&=&-2\alpha sgn(\tau_c)\underline{W_{ij}}
\end{eqnarray}

If (\ref{kirksekiz}) is inserted into (\ref{kirkyedi}), (\ref{kirkdokuz}) can be achieved:
\begin{eqnarray}\label{kirkdokuz}
\dot{\widetilde{\underline{W_{ij}}}} &=& -2\alpha \widetilde{\underline{W_{ij}}}sgn(\tau_c)-\frac{\widetilde{\underline{W_{ij}}}\sum_{i=1}^{I}\sum_{j=1}^{J} \Big(-2\alpha \underline{W_{ij}}sgn(\tau_c)\Big)}{\sum_{i=1}^{I}\sum_{j=1}^{J}\underline{W_{ij}}} \nonumber \\
&=& {} -2\alpha \widetilde{\underline{W_{ij}}}sgn(\tau_c)+\widetilde{\underline{W_{ij}}}\sum_{i=1}^{I}\sum_{j=1}^{J} \Big(2\alpha \widetilde{\underline{W_{ij}}}sgn(\tau_c)\Big) \nonumber \\
&=& {}-2\alpha \widetilde{\underline{W_{ij}}}sgn(\tau_c)+2\alpha sgn(\tau_c)\widetilde{\underline{W_{ij}}}\overbrace{\sum_{i=1}^{I}\sum_{j=1}^{J} \widetilde{\underline{W_{ij}}}}^\text{1} \nonumber \\
&=& {}-2\alpha \widetilde{\underline{W_{ij}}}sgn(\tau_c)+2\alpha \widetilde{\underline{W_{ij}}}sgn(\tau_c) \nonumber \\
&=& 0
\end{eqnarray}

Similarly, it can easily be shown that:
\begin{equation}
\dot{\widetilde{\overline{W_{ij}}}} = 0
\end{equation}

By using the following Lyapunov function, the stability condition is checked as follows:
\begin{equation}
V_c = \frac{1}{2}\tau_c^2(t)
\end{equation}
\begin{equation}\label{elliiki}
\dot{V_c} = \tau_c \dot{\tau_c} = \tau_c(\dot{\tau_n}+ \dot{\tau})
\end{equation}
\begin{eqnarray}\label{elliuc}
\tau_n&=&\frac{q(t) \sum_{i=1}^{I}\sum_{j=1}^{J}f_{ij}\underline{W_{ij}}}{\sum_{i=1}^{I}\sum_{j=1}^{J}\underline{W_{ij}}}+\frac{\big(1-q(t)\big)\sum_{i=1}^{I}\sum_{j=1}^{J}f_{ij}\overline{W_{ij}}}{\sum_{i=1}^{I}\sum_{j=1}^{J}\overline{W_{ij}}}
\nonumber \\
&=& {}q(t) \sum_{i=1}^{I}\sum_{j=1}^{J}f_{ij}\widetilde{\underline{W_{ij}}}+\big(1-q(t)\big)\sum_{i=1}^{I}\sum_{j=1}^{J}f_{ij}\widetilde{\overline{W_{ij}}}
\end{eqnarray}

The time derivative of (\ref{elliuc}) is as follows:
\begin{eqnarray}\label{ellidort}
\dot{\tau_n}=q(t)\sum_{i=1}^{I}\sum_{j=1}^{J}(\dot{f_{ij}}\widetilde{\underline{W_{ij}}} + f_{ij}\dot{\widetilde{\underline{W_{ij}}}}) && \\ \nonumber +(1-q(t))\sum_{i=1}^{I}\sum_{j=1}^{J}(\dot{f_{ij}}\widetilde{\overline{W_{ij}}} + f_{ij}\dot{\widetilde{\overline{W_{ij}}}}) && \\  \nonumber + \dot{q}(t) \sum_{i=1}^{I}\sum_{j=1}^{J}f_{ij}\widetilde{\underline{W_{ij}}}-\dot{q}(t)\sum_{i=1}^{I}\sum_{j=1}^{J}f_{ij}\widetilde{\overline{W_{ij}}}
\end{eqnarray}

If (\ref{ellidort}) is inserted into (\ref{elliiki}), the following is obtained:
\begin{eqnarray}\label{eq48}
\dot{V_c} &=& \tau_c\bigg(q(t)\sum_{i=1}^{I}\sum_{j=1}^{J}(\dot{f_{ij}}\widetilde{\underline{W_{ij}}}+ f_{ij}\cancelto{0}{\dot{\widetilde{\underline{W_{ij}}}}})  {} \nonumber \\  &&  +(1-q(t))\sum_{i=1}^{I}\sum_{j=1}^{J}(\dot{f_{ij}}\widetilde{\overline{W_{ij}}} + f_{ij}\cancelto{0}{\dot{\widetilde{\overline{W_{ij}}}}}) {}\nonumber  \\ &&+ \dot{q}(t)\sum_{i=1}^{I}\sum_{j=1}^{J}f_{ij}\big( \widetilde{\underline{W_{ij}}}-\widetilde{\overline{W_{ij}}}\big) + \dot{\tau}\bigg) \nonumber   \\
&=& \tau_c\bigg(\sum_{i=1}^{I}\sum_{j=1}^{J}\bigg(q(t)\widetilde{\underline{W_{ij}}}\dot{f_{ij}}+\big(1-q(t)\big)\widetilde{\overline{W_{ij}}}\dot{f_{ij}}\bigg) {}\nonumber  \\ &&+ \dot{q}(t)\sum_{i=1}^{I}\sum_{j=1}^{J}f_{ij}\big( \widetilde{\underline{W_{ij}}}-\widetilde{\overline{W_{ij}}}\big) +\dot{\tau}\bigg) \nonumber \\
&=& \tau_c\bigg(\sum_{i=1}^{I}\sum_{j=1}^{J}\dot{f_{ij}}\bigg(q(t)\widetilde{\underline{W_{ij}}}+\big(1-q(t)\big)\widetilde{\overline{W_{ij}}}\bigg){}\nonumber  \\ &&+ \dot{q}(t)\sum_{i=1}^{I}\sum_{j=1}^{J}f_{ij}\big( \widetilde{\underline{W_{ij}}}-\widetilde{\overline{W_{ij}}}\big)  +\dot{\tau}\bigg)
\end{eqnarray}

If Eqns. \eqref{fij} and \eqref{q_q} are inserted to the equation above, the following can be obtained:
\begin{eqnarray}
\dot{V_c}=\tau_c\Big(-2\alpha sgn{(\tau_c)}+\dot{\tau}\Big)
<\Big(-2\alpha |\tau_c|+|\tau_c| B_{\dot{\tau}}\Big)< 0
\end{eqnarray}

\section{Proof of Theorem 2}
Evaluating the time derivative of the Lyapunov function in (\ref{lyapunov}) yields:
\begin{equation}
\begin{split}
\dot{V}_{p}& =\dot{S}_{p} S_{p}=\frac{1}{k_{D}^{2} } \dot{S}_{c} S_{c} \\
& \le \frac{\left|\tau _{c} \right|}{k_{D}^{2} } \Big(-\alpha+B_{\dot{\tau}}\Big)<0, \hspace{0.24in}\forall S_c, S_p \ne 0
\end{split}
\end{equation}

\bibliography{ieee-asme-bib-file}

\begin{thebibliography}{10}
\providecommand{\url}[1]{#1}
\csname url@samestyle\endcsname
\providecommand{\newblock}{\relax}
\providecommand{\bibinfo}[2]{#2}
\providecommand{\BIBentrySTDinterwordspacing}{\spaceskip=0pt\relax}
\providecommand{\BIBentryALTinterwordstretchfactor}{4}
\providecommand{\BIBentryALTinterwordspacing}{\spaceskip=\fontdimen2\font plus
\BIBentryALTinterwordstretchfactor\fontdimen3\font minus
  \fontdimen4\font\relax}
\providecommand{\BIBforeignlanguage}[2]{{%
\expandafter\ifx\csname l@#1\endcsname\relax
\typeout{** WARNING: IEEEtran.bst: No hyphenation pattern has been}%
\typeout{** loaded for the language `#1'. Using the pattern for}%
\typeout{** the default language instead.}%
\else
\language=\csname l@#1\endcsname
\fi
#2}}
\providecommand{\BIBdecl}{\relax}
\BIBdecl

\bibitem{Kawato}
M.~Kawato, Y.~Uno, M.~Isobe, and R.~Suzuki, ``Hierarchical neural network model
  for voluntary movement with application to robotics,'' \emph{Control Systems
  Magazine, IEEE}, vol.~8, no.~2, pp. 8 --15, apr 1988.

\bibitem{ascc09}
A.~Topalov, E.~Kayacan, Y.~Oniz, and O.~Kaynak, ``Adaptive neuro-fuzzy control
  with sliding mode learning algorithm: Application to antilock braking
  system,'' in \emph{Asian Control Conference, 2009. ASCC 2009. 7th}, 27-29
  2009, pp. 784 --789.

\bibitem{gradient}
S.~Beyhan and M.~Alci, ``Extended fuzzy function model with stable learning
  methods for online system identification,'' \emph{International Journal of
  Adaptive Control and Signal Processing}, vol.~25, no.~2, pp. 168--182, 2011.

\bibitem{Astrom_Witternmark}
K.~J. Astrom and B.~Wittenmark, \emph{Adaptive Control}.\hskip 1em plus 0.5em
  minus 0.4em\relax Addison-Wesley, 1995.

\bibitem{Venelinov_1}
A.~Topalov and O.~Kaynak, ``Online learning in adaptive neurocontrol schemes
  with a sliding mode algorithm,'' \emph{IEEE Transactions on Systems, Man, and
  Cybernetics Part B: Cybernetics}, vol.~31, no.~3, pp. 445 --450, december
  2001.

\bibitem{celikyilmaz}
A.~Celikyilmaz and I.~B. Turksen, ``Uncertainty modeling of improved fuzzy
  functions with evolutionary systems,'' \emph{IEEE Transactions on Systems,
  Man, and Cybernetics Part B: Cybernetics}, vol.~38, pp. 1098 -- 1110, 2008.

\bibitem{genetik2}
S.~P. Panigrahi, S.~K. Nayak, and S.~K. Padhy, ``A genetic-based neuro-fuzzy
  controller for blind equalization of time-varying channels,''
  \emph{International Journal of Adaptive Control and Signal Processing},
  vol.~22, no.~7, pp. 705--716, 2008.

\bibitem{Parma}
G.~Parma, B.~Menezes, and A.~Braga, ``Sliding mode algorithm for training
  multilayer artificial neural networks,'' \emph{Electronics Letters}, vol.~34,
  no.~1, pp. 97 --98, 8 1998.

\bibitem{Yu}
Y.~Shuanghe, X.~Y., and Z.~M., ``Fuzzy sets and systems,'' \emph{A fuzzy neural
  network approximator with fast terminal sliding mode and its applications},
  vol. 148, no.~3, pp. 469--486, 11 2004.

\bibitem{Slotin_1}
J.~J. Slotine and W.~Li, \emph{Applied Nonlinear Control}.\hskip 1em plus 0.5em
  minus 0.4em\relax Englewood Cliffs, NJ: Prentice-Hall, 1991.

\bibitem{okyayhoca_survey1}
X.~Yu and O.~Kaynak, ``Sliding-mode control with soft computing: A survey,''
  \emph{Industrial Electronics, IEEE Transactions on}, vol.~56, no.~9, pp. 3275
  --3285, 2009.

\bibitem{okyayhoca_survey2}
O.~Kaynak, K.~Erbatur, and M.~Ertugrul, ``The fusion of computationally
  intelligent methodologies and sliding-mode control-a survey,''
  \emph{Industrial Electronics, IEEE Transactions on}, vol.~48, no.~1, pp. 4
  --17, Feb. 2001.

\bibitem{Oniz}
Y.~Oniz, E.~Kayacan, and O.~Kaynak, ``A dynamic method to forecast the wheel
  slip for antilock braking system and its experimental evaluation,''
  \emph{Systems, Man, and Cybernetics, Part B: Cybernetics, IEEE Transactions
  on}, vol.~39, no.~2, pp. 551 --560, april 2009.

\bibitem{Efe2000}
M.~O. Efe, O.~Kaynak, and X.~Yu, ``Sliding mode control of a three degrees of
  freedom anthropoid robot by driving the controller parameters to an
  equivalent regime,'' \emph{ASME Journal of Dynamic Systems, Measurement, and
  Control}, vol. 122, no.~4, pp. 632 --640, December 2000.

\bibitem{Byungkook}
B.~Yoo and W.~Ham, ``Adaptive fuzzy sliding mode control of nonlinear system,''
  \emph{Fuzzy Systems, IEEE Transactions on}, vol.~6, no.~2, pp. 315 --321, may
  1998.

\bibitem{Zadehh}
L.~Zadeh, ``Toward extended fuzzy logic-{A} first step,'' \emph{Fuzzy Sets and
  Systems}, vol. 160, pp. 3175 -- 3181, 2009.

\bibitem{Liang1}
Q.~Liang, N.~Karnik, and J.~Mendel, ``Connection admission control in {ATM}
  networks using survey-based type-2 fuzzy logic systems,'' \emph{IEEE
  Transactions on Systems, Man, and Cybernetics Part C}, vol.~38, pp. 329 --
  339, 2000.

\bibitem{chia09}
C.-F. Juang and C.-H. Hsu, ``Reinforcement interval type-2 fuzzy controller
  design by online rule generation and q-value-aided ant colony optimization,''
  \emph{Systems, Man, and Cybernetics, Part B: Cybernetics, IEEE Transactions
  on}, vol.~39, no.~6, pp. 1528 --1542, dec. 2009.

\bibitem{Juang1}
------, ``Reinforcement ant optimized fuzzy controller for mobile-robot
  wall-following control,'' \emph{Industrial Electronics, IEEE Transactions
  on}, vol.~56, no.~10, pp. 3931--3940, October 2009.

\bibitem{sepu06}
R.~Sepulveda, P.~Melin, A.~Rodriguez, A.~Mancilla, and O.~Montiel, ``Analyzing
  the effects of the footprint of uncertainty in type-2 fuzzy logic
  controllers,'' \emph{Engineering Letters}, vol.~13, pp. 138–--147, 2006.

\bibitem{Roo10}
M.~Roopaei, B.~Sahraei, T.-C. Lin, and M.-C. Chen, ``Synchronization of two
  different chaotic systems using chattering-free adaptive interval type-2
  fuzzy sliding mode control,'' in \emph{Industrial Electronics and
  Applications (ICIEA), 2010 the 5th IEEE Conference on}, jun. 2010, pp. 121
  --126.

\bibitem{MendelIE}
M.~Biglarbegian, W.~Melek, and J.~Mendel, ``Design of novel interval type-2
  fuzzy controllers for modular and reconfigurable robots: Theory and
  experiments,'' \emph{Industrial Electronics, IEEE Transactions on}, vol.~58,
  no.~4, pp. 1371 --1384, april 2011.

\bibitem{ellipsoidal}
M.~Khanesar, E.~Kayacan, M.~Teshnehlab, and O.~Kaynak, ``Analysis of the noise
  reduction property of type-2 fuzzy logic systems using a novel type-2
  membership function,'' \emph{Systems, Man, and Cybernetics, Part B:
  Cybernetics, IEEE Transactions on}, vol.~41, no.~5, pp. 1395--1406, 2011.

\bibitem{WCCI2012_australia}
E.~Kayacan, W.~Saeys, E.~Kayacan, H.~Ramon, and O.~Kaynak, ``Intelligent
  control of a tractor-implement system using type-2 fuzzy neural networks,''
  in \emph{IEEE WCCI 2012 - the 2012 IEEE World Congress on Computational
  Intelligence}, Brisbane, Australia, 10-15 June, 2012, pp. 171--178.

\bibitem{Karkee}
M.~Karkee and B.~L. Steward, ``Study of the open and closed loop
  characteristics of a tractor and a single axle towed implement system,''
  \emph{Journal of Terramechanics}, vol.~47, no.~6, pp. 379 -- 393, 2010.

\bibitem{Piyabongkarn}
D.~Piyabongkarn, R.~Rajamani, J.~A. Grogg, and J.~Y. Lew, ``Development and
  experimental evaluation of a slip angle estimator for vehicle stability
  control,'' \emph{IEEE Transactions on Control Systems Technologhy}, vol.~17,
  no.~1, pp. 78 -- 88, 2009.

\bibitem{Geng}
C.~Geng, L.~Mostefai, M.~Denai, and Y.~Hori, ``Direct yaw-moment control of an
  in-wheel-motored electric vehicle based on body slip angle fuzzy observer,''
  \emph{IEEE Transactions on Industrial Electronics}, vol.~56, no.~5, pp. 1411
  -- 1419, 2009.

\bibitem{Martins20081354}
F.~N. Martins, W.~C. Celeste, R.~Carelli, M.~Sarcinelli-Filho, and T.~F.
  Bastos-Filho, ``An adaptive dynamic controller for autonomous mobile robot
  trajectory tracking,'' \emph{Control Engineering Practice}, vol.~16, no.~11,
  pp. 1354 -- 1363, 2008.

\bibitem{Erdal_Triangular}
E.~Kayacan and O.~Kaynak, ``Sliding mode control theory-based algorithm for
  online learning in type-2 fuzzy neural networks: application to velocity
  control of an electro hydraulic servo system,'' \emph{International Journal
  of Adaptive Control and Signal Processing}, vol.~26, no.~7, pp. 645--659,
  2012.

\bibitem{Biglarbegian}
M.~B. Begian, W.~W. Melek, and J.~M. Mendel, ``Parametric design of stable
  type-2 {TSK} fuzzy systems,'' in \emph{Proceedings of the North American
  Fuzzy Information Processing Systems}, 2008, pp. 1–--6.

\bibitem{Utkin}
V.~I. Utkin, \emph{Sliding Modes in Control Optimization}.\hskip 1em plus 0.5em
  minus 0.4em\relax Springer-Verlag, 1992.

\bibitem{Slotine}
J.-J. Slotine and W.~Li, \emph{Applied Nonlinear Control}.\hskip 1em plus 0.5em
  minus 0.4em\relax Prentice Hall, 1991.

\bibitem{Brown}
R.~G. Brown and P.~Y.~C. Hwang, \emph{Introduction to Random Signals and
  Applied Kalman Filtering (4th Edition)}.\hskip 1em plus 0.5em minus
  0.4em\relax John Wiley and Sons, 2012.

\end{thebibliography}
\bibliographystyle{IEEEtran}

\begin{IEEEbiography}[{\includegraphics[width=1in,height=1.25in,clip,keepaspectratio]{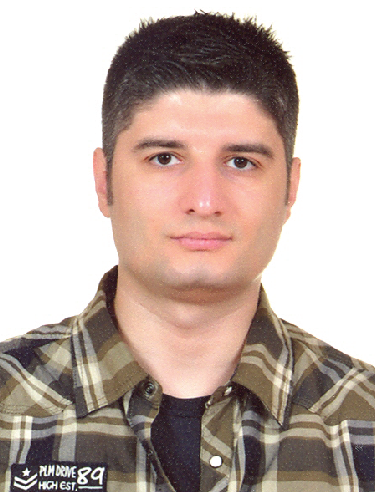}}]{Erdal Kayacan} (S\textquoteright 06-SM\textquoteright 12)  was born in Istanbul, Turkey on January 7, 1980. He received a B.Sc. degree in electrical engineering from in 2003 from Istanbul Technical University in Istanbul, Turkey as well as a M.Sc. degree in systems and control engineering in 2006 from Bogazici University in Istanbul, Turkey. In September 2011, he received a Ph.D. degree in electrical and electronic engineering at Bogazici University in Istanbul, Turkey. He is currently pursuing his post-doctoral degree at University of Leuven (KU Leuven) in the division of mechatronics, biostatistics and sensors (MeBioS). His research areas are robotics, mechatronics, soft computing methods, sliding mode control and model predictive control.

Dr. Kayacan is active in the IEEE CIS Student Activities Subcommittee, the IEEE CIS Social Media Subcommittee, the IEEE CIS GOLD Subcommittee and the IEEE SMC Technical Committee on Grey Systems. He has been serving as an editor in Journal on Automation and Control Engineering (JACE) and editorial advisory board in Grey Systems Theory and Application.
\end{IEEEbiography}

\begin{IEEEbiography}[{\includegraphics[width=1in,height=1.25in,clip,keepaspectratio]{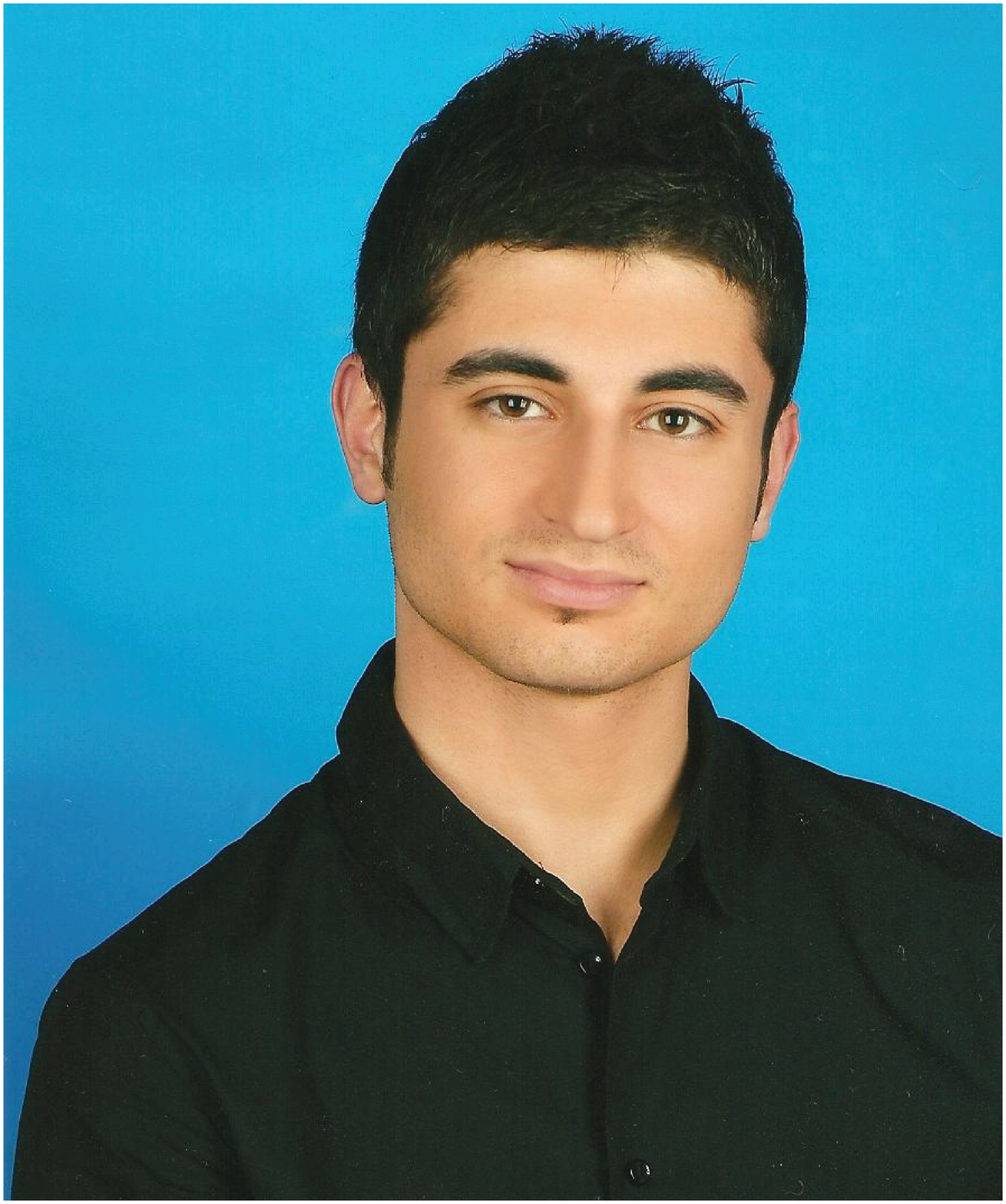}}]{Erkan Kayacan} (S\textquoteright 12) was born in Istanbul, Turkey, on April 17, 1985. He received the B.Sc. and the M.Sc. degrees in mechanical engineering from Istanbul Technical University, Istanbul, in 2008 and 2010, respectively. He is a PhD student and research assistant at University of Leuven (KU Leuven) in the division of mechatronics, biostatistics and sensors (MeBioS). His research interests include model predictive control, moving horizon estimation, distributed and decentralized control, intelligent control and vehicle dynamics.

He is active in the IEEE CIS GOLD Subcommittee and the IEEE Webinars subcommittee.
\end{IEEEbiography}

\begin{IEEEbiography}[{\includegraphics[width=1in,height=1.25in,clip,keepaspectratio]{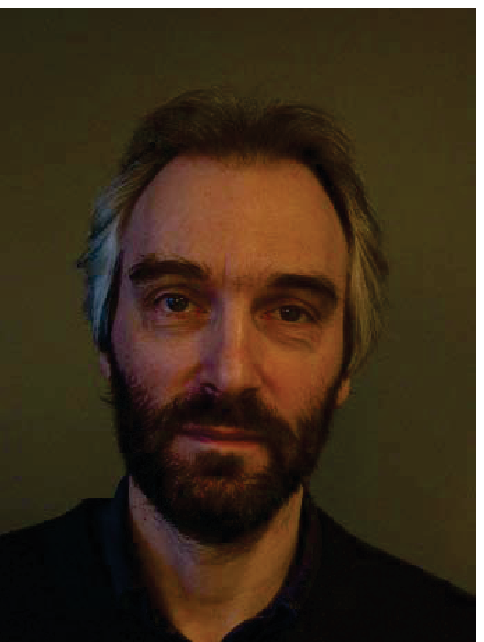}}]{Herman Ramon} graduated as an agricultural engineer from Gent University. In 1993 he obtained a Ph.D. in applied biological sciences at the Katholieke Universiteit Leuven. He is currently Professor at the Faculty of Agricultural and Applied Biological Sciences of the Katholieke Universiteit Leuven, lecturing on agricultural machinery and mechatronic systems for agricultural machinery. He has a strong research interest in precision technologies and advanced mechatronic systems for processes involved in the production chain of food and nonfood materials, from the field to the end user.He is author or co-author of more than 40 papers.
\end{IEEEbiography}

\begin{IEEEbiography}[{\includegraphics[width=1in,height=1.25in,clip,keepaspectratio]{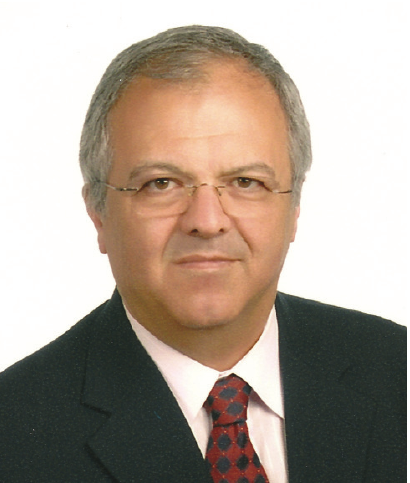}}]{Okyay Kaynak}  (M\textquoteright 80-SM\textquoteright 90-F\textquoteright 03) received the B.Sc. degree with first class honors and Ph.D. degrees in Electronic and Electrical Engineering from the University of Birmingham, UK, in 1969 and 1972 respectively.

From 1972 to 1979, he held various positions within the industry. In 1979, he joined the Department of Electrical and Electronics Engineering, Bogazici University, Istanbul, Turkey, where he is presently a full professor, holding the UNESCO Chair on Mechatronics. He has hold long-term (near to or more than a year) Visiting Professor / Scholar positions at various institutions in Japan, Germany, U.S. and Singapore. His current research interests are in the fields of intelligent control and mechatronics. He has authored three books and edited five and authored or coauthored almost 400 papers that have appeared in various journals and conference proceedings. Currently, he is the Editor in Chief of IEEE/ASME Trans. on Mechatronics.

Dr. Kaynak is a fellow of IEEE. He has served on many committees of IEEE and was the president of IEEE Industrial Electronics Society during 2002-2003.
\end{IEEEbiography}

\begin{IEEEbiography}[{\includegraphics[width=1in,height=1.25in,clip,keepaspectratio]{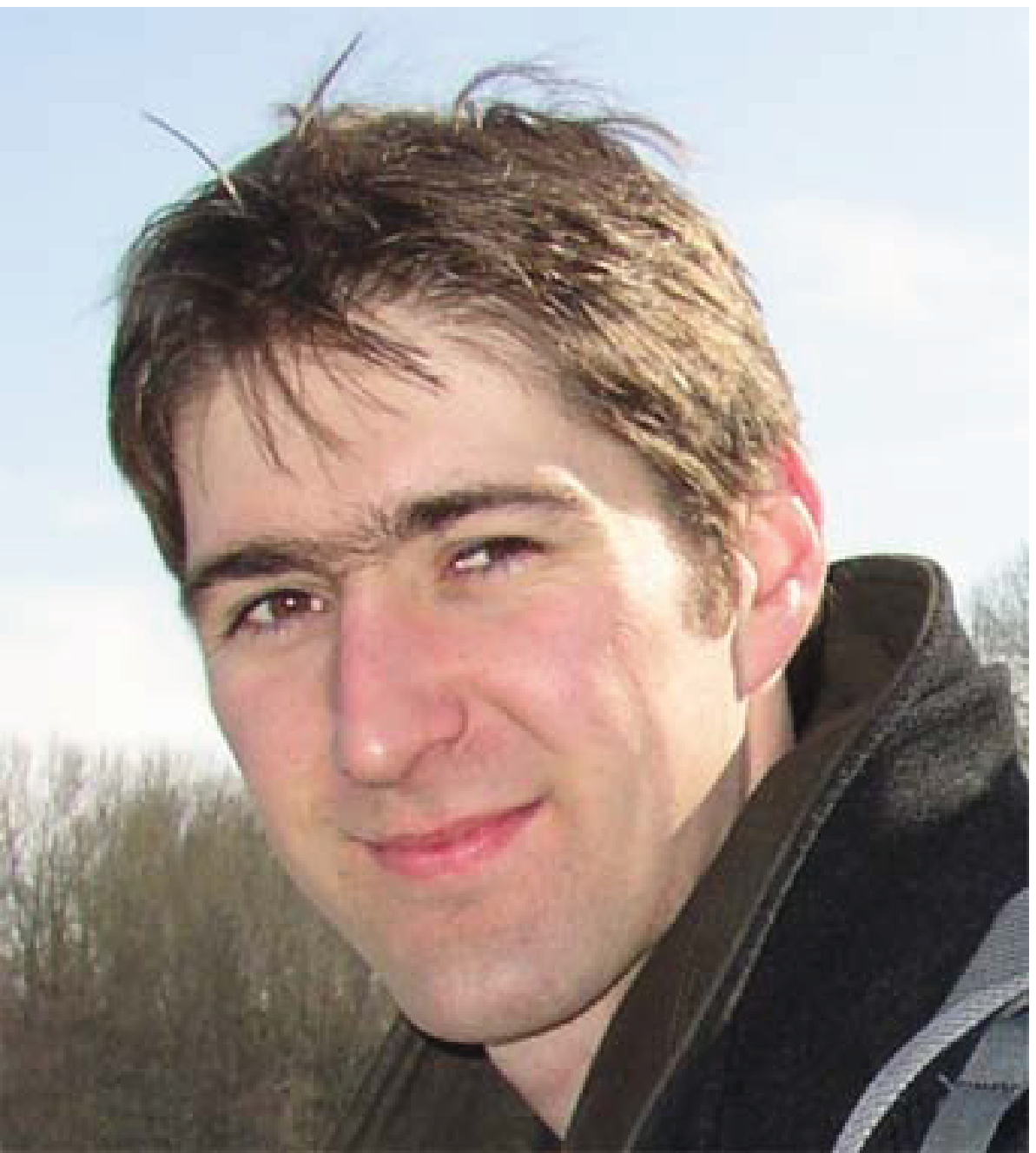}}]{Wouter Saeys} is currently Assistant Professor in Biosystems Engineering at the Department of Biosystems of the University of Leuven in Belgium. He obtained his Ph.D. at the same institute and was a visiting postdoc at the School for Chemical Engineering and Advanced Materials of the University of Newcastle upon Tyne, UK and at the Norwegian Food Research Institute - Nofima Mat in Norway. His main research interests are optical sensing, process monitoring and control with applications in food and agriculture. He is author of 50 articles (ISI) and member of the editorial board of Biosystems Engineering.
\end{IEEEbiography}

\end{document}